\documentclass[twoside]{article}

\usepackage[accepted]{aistats2020}
\usepackage[utf8]{inputenc} % allow utf-8 input
\usepackage[T1]{fontenc}    % use 8-bit T1 fonts
\usepackage{lmodern}
\usepackage{hyperref}       % hyperlinks
\usepackage{url}            % simple URL typesetting
\usepackage{booktabs}       % professional-quality tables
\usepackage{amsfonts}       % blackboard math symbols
\usepackage{nicefrac}       % compact symbols for 1/2, etc.
\usepackage{microtype}      % microtypography
\usepackage{floatrow}

\newcommand{\RR}{\mathbb{R}} %real numbers
\newcommand{\xx}{\mathbf{x}} 

\newcommand{\x}{\mathbf{x}} 
\newcommand{\y}{\mathbf{y}} 
\newcommand{\s}{\mathbf{s}} 
\newcommand{\EE}{\mathbb{E}}
\newcommand{\GP}{\mathcal{GP}}
\newcommand{\NN}{\mathcal{N}}
\newcommand{\mut}{\widetilde{\mu}}
\newcommand{\St}{\widetilde{\Sigma}}
\newcommand{\UU}{\mathbf{U}}
\newcommand{\mm}{\mathbf{m}}
\newcommand{\Sb}{\mathbf{S}}
\newcommand{\z}{\mathbf{z}}

\newcommand{\cp}[2]{p\left(#1 \, | \, #2\right)}
\newcommand{\KL}[2]{\text{KL}[#1 \, || \, #2]}
\newcommand{\given}{\: | \:}

\newcommand{\ZZ}{\mathbf{Z}}
\newcommand{\zz}{\mathbf{z}}
\newcommand{\Kxz}{\mathbf{K}_{(s, t), \z}}
\newcommand{\Kzx}{\mathbf{K}_{\z, (s, t)}}
\newcommand{\Kzz}{\mathbf{K}_{\mathbf{zz}}}

\newcommand{\Kab}{\mathbf{K}_{\mathbf{ab}}}

\usepackage[font=footnotesize]{subcaption} % To use subfigures with subcaptions
\usepackage[font=small]{caption}
\usepackage{enumerate}
\usepackage{enumitem}
\usepackage{amsmath}
\usepackage{amssymb}
\usepackage{adjustbox}

\usepackage{enumitem}

\usepackage[colorinlistoftodos,prependcaption,textsize=tiny]{todonotes}

\usepackage{comment}

\newcommand{\nonparametric}{nonparametric}

\usepackage{rotating}

\usepackage{tabularx}
\newcolumntype{Y}{>{\centering\arraybackslash}X}

\newcommand\blfootnote[1]{%
  \begingroup
  \renewcommand\thefootnote{}\footnote{#1}%
  \addtocounter{footnote}{-1}%
  \endgroup
}

\renewcommand\floatpagefraction{.9}
\renewcommand\dblfloatpagefraction{.9} % for two column documents
\renewcommand\topfraction{.9}
\renewcommand\dbltopfraction{.9} % for two column documents
\renewcommand\bottomfraction{.9}
\renewcommand\textfraction{.1}   
\begin{document}


\runningauthor{Ustyuzhaninov, Kazlauskaite, Ek, Campbell}

\twocolumn[

\aistatstitle{Monotonic Gaussian Process Flows}%: Supplementary material }

% \aistatsauthor{ Ivan Ustyuzhaninov$^{* 1}$ \And  Ieva Kazlauskaite$^{* 2}$ \And  Carl Henrik Ek$^3$ \And  Neill D. F. Campbell$^{2,4}$ }

% \aistatsaddress{ \\ \vspace{-0.4cm} $^1$University of T\"ubingen \And \\  $^2$University of Bath \And  \\ $^3$University of Bristol \And \\ $^4$Royal Society} ]

\aistatsauthor{ Ivan Ustyuzhaninov$^{*}$ \And  Ieva Kazlauskaite$^{*}$ \And  Carl Henrik Ek \And  Neill D. F. Campbell }

\aistatsaddress{ University of T\"ubingen \And University of Bath, \\ Electronic Arts \And University of Bristol \And University of Bath, \\ Royal Society} ]

% \author{
%   Ivan Ustyuzhaninov *\\
% %  Centre for Integrative Neuroscience\\
%   University of T\"ubingen\\
%   Germany \\
% %  \texttt{ivan.ustyuzhaninov@uni-tuebingen.de}
%   \And
%   Ieva Kazlauskaite * \\
%   University of Bath, UK\\
%   Electronic Arts \\
% %   \texttt{i.kazlauskaite@bath.ac.uk} \\
%   \AND
%   Carl Henrik Ek \\
%   University of Bristol, UK \\
%   \And
%   Neill D. F. Campbell \\
%   University of Bath, UK \\
%   Royal Society \\
% %   \texttt{n.campbell@bath.ac.uk}
% } 

\begin{abstract}
We propose a new framework for imposing monotonicity constraints in a Bayesian \nonparametric{} setting based on numerical solutions of stochastic differential equations. We derive a \nonparametric{} model of monotonic functions that allows for interpretable priors and principled quantification of hierarchical uncertainty. We demonstrate the efficacy of the proposed model by providing competitive results to other probabilistic monotonic models on a number of benchmark functions. In addition, we consider the utility of a monotonic random process as a part of a hierarchical probabilistic model; we examine the task of temporal alignment of time-series data where it is beneficial to use a monotonic random process in order to preserve the uncertainty in the temporal warpings. \blfootnote{* Equal contribution}
\end{abstract}

% !TEX root = ../neurips_2019.tex
\section{INTRODUCTION}
\label{intro}
Monotonic regression is a task of inferring the relationship between a dependent variable $y$ and an independent variable $x$ when it is known that the relationship $y = f (x)$ is monotonic. 
Monotonic functions (and monotonic random processes) have previously been studied in areas as diverse as physical sciences for estimating the temperature of a cannon barrel over time~\cite{Lavine:1995}, marine biology for surveying of fauna on the seabed of the Great Barrier Reef~\cite{Hall:2001}, geology for chronology of sediment samples~\cite{Haslett:2008}, public health for relating obesity and body fat~\cite{Dette:2006}, sociology for relating education, work experience and salary~\cite{Dette:2006}, design of computer networking systems~\cite{Golchi:2015}, economics for estimating personal income~\cite{Canini:2016}, insurance for predicting mortality rates ~\cite{Durot:2018}, biology of establishing the diagnostic value of bio-markers for Alzheimer’s disease 
and for trajectory estimation in brain imaging~\cite{Lorenzi:2019, Nader:2019}, meteorology for estimation of wind-induced under-catch of winter precipitation~\cite{Kim:2018} and others.

Monotonicity also appears in the more general context of hierarchical models where we want to transform a (simple and typically stationary) input distribution to a (complicated and non-stationary) data distribution.
More specifically, monotonicity constraints have been used in hierarchical models with warped inputs, for example, in Bayesian optimisation of non-stationary functions~\cite{Snoek:2014} and in mixed effects models for temporal warps of time-series data~\cite{Kaiser:2018, Kazlauskaite:2019, Raket:2016}.

Extensive study by the statistics~\cite{Ramsay:1988, Sill:1997} and machine learning communities~\cite{Riihimaki:2010, Andersen:2018} has resulted in a variety of frameworks. While many traditional approaches use constrained parametric splines, they are not sufficiently expressive and, typically, do not include prior beliefs about the characteristics of the underlying function (such as smoothness).
Consequently, many contemporary methods consider monotonicity in the context of continuous random processes, mostly based on Gaussian processes (GPs)~\cite{Rasmussen:2005}. 
As a \nonparametric{} Bayesian model, a GP is an attractive foundation on which to build flexible and theoretically sound models with well-calibrated estimates of uncertainty and automatic complexity control. 
However, imposing monotonicity constraints on a GP has proven to be problematic~\cite{Lin:2014, Riihimaki:2010} as it requires both formulating a prior that is monotonic as well as constraining the (predictive) posterior to be monotonic. 
This is particularly challenging as monotonicity is a global property, implying that the function values are correlated for all inputs, irrespective of the lengthscale of the covariance~\cite{Andersen:2018}.

In this work we propose a novel \nonparametric{} Bayesian model of monotonic functions that is based on the recent work on differential equations (DEs). At the heart of such models is the idea of approximating the derivatives of a function rather than studying the function directly. DE models have gained a lot of popularity recently and they have been successfully applied in conjunction with both neural networks~\cite{Chen:2018} and GPs~\cite{Heinonen:2018, Yildiz:2018a, Yildiz:2018}. We consider a recently proposed framework, called differential GP flows~\cite{Hedge:2019}, that performs classification and regression by learning a stochastic differential equation (SDE) transformation of the input space. It admits an expressive yet computationally convenient parametrisation using GPs. 

Utilising the uniqueness theorem for the solutions of SDEs~\cite{Oksendal:1992}, we formulate a novel stochastic random process that is guaranteed to be monotonic.
% making this model satisfy the properties outlined in a previous paragraph. 
We show that, unlike some of the previous work on monotonic random processes, the proposed approach is guaranteed to lead to monotonic samples from the model (defined as a flow field), and it performs competitively on a set of regression benchmarks.

Furthermore, we study an illustrative example of a hierarchical, two-layer model where the first layer corresponds to a smooth monotonic warping of time and the second layer corresponds to a sequence of time-series observations.
The overall goal in such a problem is to learn the warpings of the inputs such that the unwarped versions of the sequences are temporally aligned.
While such models typically rely on parametric transformations for the temporal warpings~\cite{Kazlauskaite:2019}, we show how the estimation of uncertainty leads to a model that is more informative and more interpretable than the previous approach. 
To achieve this, we further make use of the recent advances in variational inference for deep GPs~\cite{Ustyuzhaninov:2019} to capture the compositional uncertainty present in the hierarchical model.

\section{RELATED WORK}
\label{subsec:related-work}

\paragraph{Splines} Many classical approaches to monotonic regression rely on spline smoothing: given a basis of monotone spline functions, the underlying function is approximated using a non-negative linear combination of these basis functions and the monotonicity constraints are satisfied in the entire domain~\cite{Wahba:1978} by construction. For example, Ramsey~\cite{Ramsay:1998} considers a family of functions defined by the differential equation $D^2f = \omega Df$ which contains the strictly monotone twice differentiable functions, and approximates $\omega$ using a basis of M-splines and I-splines. Shively~\emph{et al.}~\cite{Shively:2009} consider a finite approximation using quadratic splines and a set of constraints on the coefficients that ensure isotonicity at the interpolation knots. The use of piecewise linear splines was explored by Haslett and Parnell~\cite{Haslett:2008} who use additive i.i.d.~gamma increments and a Poisson process to locate the interpolation knots; this leads to a process with a random number of piecewise linear segments of random length, both of which are marginalised analytically. Further examples of spline based approaches rely on cubic splines~\cite{Wolberg:2002}, mixtures of cumulative distribution functions~\cite{Bornkamp:2009} and an approximation of the unknown regression function using Bernstein polynomials~\cite{Curtis:2011}. \vspace{-5pt}

\paragraph{Gaussian process} A GP is a stochastic process which is fully specified by its mean function, $\mathbf{\mu(x)}$, and its covariance function, $k(\mathbf{x}, \mathbf{x}')$, such that any finite set of random variables have a joint Gaussian distribution~\cite{Rasmussen:2005}.
GPs provide a robust method for modeling non-linear functions in a Bayesian \nonparametric{} framework; ordinarily one considers a GP prior over the function and combines it with a suitable likelihood to derive a posterior estimate for the function given data. The \nonparametric{} nature of the GP means that, unlike the parametric counterparts, it adapts to the complexity of the data. \vspace{-5pt}

\paragraph{Monotonic Gaussian processes} A common approach is to ensure that the monotonicity constraints are satisfied at a finite number of input points. 
For example, Da Veiga and Marrel~\cite{DaVeiga:2012} use a truncated multi-normal distribution and an approximation of conditional expectations at discrete locations, while Maatouk~\cite{Maatouk:2017} and Lopez-Lopera~\emph{et al.}~\cite{Lopez-Lopera:2019} proposed finite-dimensional approximations based on deterministic basis functions evaluated at a set of knots. 
Another popular approach proposed by Riihim\"aki and Vehtari~\cite{Riihimaki:2010} is based on including the derivatives information at a number of input locations by forcing the derivative process to be positive at these locations. 
Extensions to this approach include both adapting to new application domains~\cite{Golchi:2015, Lorenzi:2019, Siivola:2016} and proposing new inference schemes~\cite{Golchi:2015}. However, these approaches do not guarantee monotonicity as they impose constraints at a finite number of points only.  
Lin and Dunson~\cite{Lin:2014} propose another GP based approach that relies on projecting sample paths from a GP to the space of monotone functions using pooled adjacent violators algorithm which does not impose smoothness. 
Furthermore, the projection operation complicates the inference of the parameters of the GP and produces distorted credible intervals. 
Lenk and Choi~\cite{Lenk:2017} design shape restricted functions by enforcing that the derivatives of the functions are squared Gaussian processes and approximating the GP using a series expansion with the Karhunen-Lo\`eve representation and numerical integration. 
Andersen~\emph{et al.}~\cite{Andersen:2018} follow a similar approach, in which the derivatives of the functions are assumed to be compositions of a GP and a non-negative function; in the following we refer to this method as transformed GP.

% construct monotonic functions using a differential equation formulation where the solution to the DE is constructed using a composition of a GP and a non-negative function; we refer to this method as transformed GP.

% !TEX root = ../neurips_2019.tex
\section{BACKGROUND}
\label{sec:background}

We now discuss the SDE framework we build upon for our monotonic random process.
Any random process can be defined through its finite-dimensional distribution~\cite{Oksendal:1992}. This implies that modelling the observations $\{f(x_n)\}_{n=1}^{N}$ with trajectories of such a process requires their definition through the finite-dimensional joint distributions $p(f(x_1), \ldots, f(x_N))$. 
Constraining the functions %$\{f(x)\}$ 
to be monotonic necessitates choosing a family of joint probability 
distributions that satisfies the monotonicity constraint:
\begin{equation}
\begin{aligned}
    p(f(x_1), \ldots, f(x_N)) &= 0, \\ \text{unless} \: f(x_1) &\leq \ldots \leq f(x_N).
\end{aligned}\label{eq:monotonicity-constraint}\tag{MC}
\end{equation}
This could be achieved by truncating a standard joint distribution (\emph{e.g.}~Gaussian) but inference in such models is computationally challenging~\cite{Maatouk:2017}. Another approach is to define a random process to have monotone trajectories by construction (\emph{e.g.}~Compound Poisson process) but this often requires making simplifying assumptions on the trajectories (and therefore on $\{f(x)\}$). In contrast, we use solutions of SDEs to define a random process with monotonic trajectories by construction while avoiding strong simplifying assumptions.

\subsection{Gaussian process flows}
\label{subsec:background-gp-flows}

\paragraph{SDE solutions} Our model builds on the general framework for modelling functions as SDE solutions introduced in \cite{Hedge:2019}. Consider the following SDE:
\begin{equation}
\begin{aligned}
    dS(t, \omega; x ) &= \mu\left(S(t, \omega ; x ), t\right)\, dt \\
    &+ \sqrt{\sigma \left(S(t, \omega ; x), t \right)} \, dW\!(t, \omega)
    \label{eq:sde}
\end{aligned}
\end{equation} where $W\!(t, \omega)$ is the Wiener process.
The solution of this SDE is a stochastic process $S(t,\omega ; x)$ which is a function of three arguments: the time $t$, the initial value $x$ at time $t = 0$, and the element $\omega \in \Omega$ of the underlying sample space $\Omega$.\footnote{Typically, the dependencies on $x$ and $\omega$ are omitted, denoting the stochastic process as $S_t$, however, these dependencies are crucial for our construction of the monotonic flow model, thus we explicitly keep them in the notation.}

For a fixed time $t = T$, the corresponding SDE solution $S(T,\omega ; x)$ is a random variable that depends on the initial condition $x$.
Therefore, there exists a mapping of an arbitrary initial condition to this solution at time $T$: $x \mapsto S(T,\omega ; x)$ and the distribution of the SDE solutions induces a distribution over such mappings (similar to GPs, for example). 
The family of such distributions is parametrised by functions $\mu \left(S(t, \omega ; x), t \right)$ (drift) and $\sigma \left(S(t, \omega ; x), t \right)$ (diffusion), which are defined in \cite{Hedge:2019} using a sparse Gaussian process~\cite{Titsias:2009}.

\paragraph{Flow GP}
Consider a zero-mean, single-output GP $\protect{g \sim \GP(0, \,k(\cdot,\cdot))}$, which is a function of two arguments: a space variable $s$ and time $t$.
% with $g: \RR^2 \to \RR$ since $g$ is a function of both $x$ and time $t$.
We specify the GP via a set of $M$ inducing outputs $\UU = \{U_m\}_{m=1}^M$, $U_m \in \RR$, corresponding to inducing input locations $\ZZ = \{\zz_m\}_{m=1}^M$, $\zz_m \in (\{s\} \times \{t\}) = \RR^{2}$, similarly to~\cite{Titsias:2009}. 
The predictive posterior distribution of such a GP evaluated at a spatio-temporal point $(s, t)$ is as follows: 
\begin{equation}
    \begin{aligned}
        p(g(s, t) &\mid \UU, \ZZ)  \sim \NN( \mut(s, t), \St(s, t) ) \\
        \mut(s, t) & = \Kxz \, \Kzz^{-1} \, \UU, \\ %\quad \text{where } \Kab := k(\mathbf{a}, \mathbf{b}),\\ 
        \St(\xx, t) & = k((s, t), (s, t)) - \Kxz \, \Kzz^{-1} \, \Kzx,
        \label{eq:gp-posterior}
    \end{aligned}
\end{equation} where the covariance matrix  $\Kab := k(\mathbf{a}, \mathbf{b})$.
We define the SDE drift and diffusion functions to be $\mu(S(t, \omega; x ), t) := \mut(S(t, \omega; x ), t)$ and $\sigma(S(t, \omega; x ), t) := \St(S(t, \omega; x ), t)$ implying that \eqref{eq:sde} is completely defined by the GP $g$ and its set of inducing points $\{\UU, \ZZ\}$. Similarly to~\cite{Hedge:2019}, the joint density of a single path then is (neglecting $\ZZ$ for clarity):
\begin{equation}
\begin{aligned}
p(y, S(T, \omega; \, &x ), g, \UU) = \underbrace{p(y \given S(T, \omega; x ))}_{\text {likelihood }}  \\ &\underbrace{p\left(S(T, \omega; x ) \given g\right)}_{\text {SDE }} \: \underbrace{p\left(g \given \mathbf{U}\right) p\left(\mathbf{U}\right)}_{\text {GP prior of } g(s, t)}.
\end{aligned}
\end{equation}

\paragraph{Inference} Inferring $\UU$ is intractable in closed form, hence the posterior of $\UU$ is approximated by a variational distribution $q(\UU) \sim \NN(\mm, \Sb)$, the parameters of which (and the inducing inputs $\ZZ$) are optimised by maximising the marginal likelihood lower bound $\mathcal{L}$:
\begin{equation}
\begin{aligned}
    \log p&(\mathcal{D}) \geq \mathcal{L} := -\,\KL{\,q(\UU)} {p(\UU)\,}
    \\ + &\,\EE_{q(\UU)} \EE_{\cp{S(T, \omega; x )}{\UU}} [\,\log \cp{y}{S(T, \omega; x )}\,]  .
    \label{eq:flow-lower-bound}
\end{aligned}
\end{equation}

The expectation $\EE_{\cp{S(T, \omega; \x )}{\UU}}$ is approximated by sampling the numerical approximations of the SDE solutions. This is particularly convenient to do with $\mu \left(S(t, \omega; x ), t \right)$ and $\sigma \left(S(t, \omega; x ), t \right)$ defined as parameters of a GP posterior as sampling such an SDE solution only requires generating samples from the posterior of the GP given the inducing points $\UU$ (see~\cite{Hedge:2019} for details). The first term in \eqref{eq:flow-lower-bound} is a KL divergence between two Gaussian distributions available in closed form.
% !TEX root = ../neurips_2019.tex
\section{MONOTONIC GAUSSIAN PROCESS FLOW}
\label{sec:monotonic-flow}

We now describe our proposed random process with monotonic trajectories.
Assuming $N$ one-dimensional initial conditions (denoted jointly as $\protect{\x = (x_1, \ldots, x_N) \in \RR^N}$), we use the SDE solution mapping $\protect{\x \mapsto S(T,\omega; \x )} := (S(T,\omega; x_1), \ldots, S(T,\omega; x_N))$ as our model of monotonic function. 
We begin with an intuitive discussion of why $S(T,\omega; \x )$ is a monotonic function of $\x$ using a fluid flow field analogy. % before providing a formal argument.
\begin{enumerate}[leftmargin=1.5em,itemsep=0pt,topsep=-3pt]
    \item A general ordinary smooth DE $du(t) = \phi(u) dt$ may be thought of as a fluid flow field.
    Its solutions $u(t, x_1), \ldots, u(t, x_n)$ corresponding to the initial values $x_1, \ldots, x_n$ are trajectories or streams of particles in this field starting at these initial values.
    A fundamental property of such flows is that one can never cross the streams of the flow field.\footnote{Also an important safety tip to avoid total protonic reversal~\cite{ghostbusters1984}.}
    Therefore, if particles are evolved simultaneously under a flow field their ordering cannot be permuted; this gives rise to a monotonicity constraint.
    \item A stochastic differential equation, however, introduces random perturbations into the flow field so particles evolving \emph{independently} could jump across flow lines and change their ordering. However, a single, coherent draw from the SDE (corresponding to an individual realisation of the paths $W\!(\cdot, \omega)$) will always produce a valid flow field (the flow field will simply change between draws). Thus, particles evolving jointly under a single draw will still evolve under a valid flow field and therefore never permute.
\end{enumerate}

\subsection{SDE solutions are monotonic functions of initial values}
\label{subsec:sde-monotonic-proof}
The joint distribution $p\left( S(T, \omega; x_1), \ldots, S(T, \omega; x_N) \right)$ of solutions of the SDE in~\eqref{eq:sde} with initial values $\protect{x_1 \leq \ldots \leq x_N}$ 
satisfies~\eqref{eq:monotonicity-constraint}. 

This follows from a general result that SDE solutions $S(t,\omega; x )$ are unique and continuous under certain regularity assumptions for any initial value $x$ (see, for example, Theorem 5.2.1 in~\cite{Oksendal:1992}). 
Specifically, a random variable $S(t,\omega; x )$ is a unique and continuous function of $t$ for any element of the sample space $\omega \in \Omega$.
Using this result we conclude that if we have two initial conditions $x$ and $x'$ such that $x \leq x'$, the corresponding solutions at some time $T$ also obey this ordering, \emph{i.e.} $S(T,\omega; x ) \leq S(T,\omega; x' )$ for $\omega \in \Omega$. 
Indeed, were that not the case, the continuity of $S(t,\omega; x )$ as a function of $t$ implies that there exists some $0 \leq t_c \leq T$ such that $S(t_c,\omega; x ) = S(t_c,\omega; x' )$ (\emph{i.e.}~the trajectories corresponding to initial values $x$ and $x'$ cross), resulting in two different solutions of the SDE for the initial condition $x_c := S(t_c,\omega; x ) = S(t_c,\omega; x' )$ (namely $S(T - t_c, \omega; x_c) = S(T,\omega; x )$ and $S(T - t_c, \omega; x_c) = S(T,\omega; x' )$), violating the uniqueness result.

The above argument assumes a fixed flow field (defined by the drift and the diffusion functions) and a fixed Wiener realisation (corresponding to $W\!(\cdot, \omega)$); it implies that individual solutions (\emph{i.e.}~solutions to a single draw) of the SDEs at a fixed time $T$, $S(T,\omega; x )$, are monotonic functions of the initial conditions, and hence define a random process with monotonic trajectories. 
The actual prior distribution of such trajectories depends on the exact form of the functions $\mu \left(S(t,\omega; x ), t \right)$ and $\sigma \left(S(t,\omega; x ), t \right)$ in \eqref{eq:sde} (\emph{e.g.}~if $\sigma(S(t,\omega; x ), t) = 0$, the SDE is an ordinary DE and $S(T,\omega; x )$ is a deterministic function of $\x$ independent of $\omega$, meaning that the prior distribution consists of a single monotonic function). 
Prior distributions over $\mu \left(S(t,\omega; \x ), t \right)$ and $\sigma \left(S(t,\omega; \x ), t \right)$ thus induces priors over the monotonic functions $S(T,\omega; \x )$, and inference in this model consists of computing the posterior distribution of these functions conditioned on the observed noisy sample from a monotonic function. For details of the numerical solution of the SDE, see supplementary material~\ref{sec:numerical_solution}.

\subsection{Notable differences to~Hedge~et~al.}%\cite{Hedge:2019}}
\begin{enumerate}
    \item In~\cite{Hedge:2019}, a regular GP is placed on top of the SDE solutions $S(T, \omega; \x )$, so that $\cp{\y}{S(T, \omega; \x )}$ is a GP with a Gaussian likelihood in \eqref{eq:flow-lower-bound}. In contrast, since we are modelling monotonic functions and $S(T, \omega; \x )$ are monotonic functions of $\xx$, we define $\cp{Y}{S(T, \omega; \x )}$ to be directly the likelihood
    \begin{equation}
        \cp{\y}{S(T, \omega; \x )} = \NN(\y \, | \, S(T, \omega; \x ), \sigma^2 I),    
    \end{equation}
    where $\y$ is a vector of observations sampled from an underlying unknown monotonic function $f(\x)$.
    \item The argument in this section assumes a fixed flow field (defined by the drift and the diffusion functions) and a fixed Wiener realisation (denoted by $\omega$). 
    Thus, a critical difference in our inference procedure is that at every iteration of the numerical SDE solver, we jointly sample the increments $\Delta \x$ in the flow field using \eqref{eq:gp-posterior}.
    This ensures that they are taken from the same instantaneous realisation of the stochastic flow field and hence the monotonicity constraint is satisfied.
\end{enumerate}

% \input{includes/two-layers.tex}
% !TEX root = ../neurips_2019.tex
\section{EXPERIMENTS} \label{sec:experiments}

First, we test the monotonic flow model on the task of estimating monotonic curves from noisy observations (in high and low data regimes) before investigating the quantification of uncertainty.

\paragraph{Regression}
\label{sec:regression}%
We use a set of $6$ benchmark functions from previous studies~\cite{Lin:2014, Maatouk:2017, Shively:2009}.
Three examples of the functions are shown in Fig.~\ref{fig:monotonic-regression}; the exact equations are in the Supplement~\ref{supp:regression_functions}. The training data is generated by evaluating these functions at $N$ equally spaced points and adding i.i.d.~Gaussian noise $\varepsilon_n \sim \NN(0, 1)$. 
We note that many real-life datasets that benefit from monotonicity constraints have similar trends and high levels of noise (\emph{e.g.}~\cite{Haslett:2008, Curtis:2011, Kim:2018}). Following the literature, we used the root-mean-square-error (RMSE)
to evaluate performance.

\paragraph{100 data points} Table~\ref{table:evals} in the Supplement~\ref{supp:quantitative-results} provides the results obtained by fitting different monotonic models to data sets containing $N = 100$ points. As baselines we include: GPs with monotonicity information~\cite{Riihimaki:2010}\footnote{Implementation available from \url{https://research.cs.aalto.fi/pml/software/gpstuff/}.}, transformed GPs~\cite{Andersen:2018}\footnote{Implementation provided in personal communications.}, and other results reported in the literature. We report the RMSE means and the SD from 20 trial runs with different random noise samples and show example fits in the bottom row of Fig.~\ref{fig:monotonic-regression}.
This figure contains the means of the predicted curves from $10$ trials with the best parameter values (each trial contains a different sample of standard Gaussian random noise). 
We plot samples as opposed to the mean and the SD as, due to the monotonicity constraint, samples are more informative than sample statistics. The parameter values we cross-validated over are detailed in the Supplement~\ref{supp:regression-parameters}.

Overall, our method performs very competitively, achieving the best results on 3 functions and being within a standard deviation of the best result on all others. 
We note that the training data contains a lot of observational noise (see Fig.~\ref{fig:monotonic-regression}), thus using prior monotonicity assumptions significantly improves results over a regular GP.

\paragraph{15 data points} In Table~\ref{tab:15pts} in the Supplement~\ref{supp:quantitative-results} and Fig.~\ref{fig:monotonic-regression} (top row) we provide a comparison of the flow and the transformed GP in a setting when only $N = 15$ data points are available. Our fully \nonparametric{} model is able to recover the structure in the data significantly better than the Transformed GP which usually reverts to a nearly linear fit on all functions. 
This might be explained by the fact that the Transformed GP is a parametric approximation of a monotonic GP, and the more parameters included, the larger the variety of the functions it can model. However, estimating a large (w.r.t.~dataset size) number of parameters is challenging given a small set of noisy observations. 
The monotonic flow tends to underestimate the value of the function on the left side of the domain and overestimate the value on the right. The mean of our prior of the monotonic flow with a stationary flow GP kernel is an identity function, so given a small set of noisy observations, the predictive posterior mean quickly reverts to the prior distribution near the edges of the data interval.
\begin{figure*}[t]
\centering
    \begin{subfigure}[h]{0.31\textwidth}
        \includegraphics[width=\textwidth]{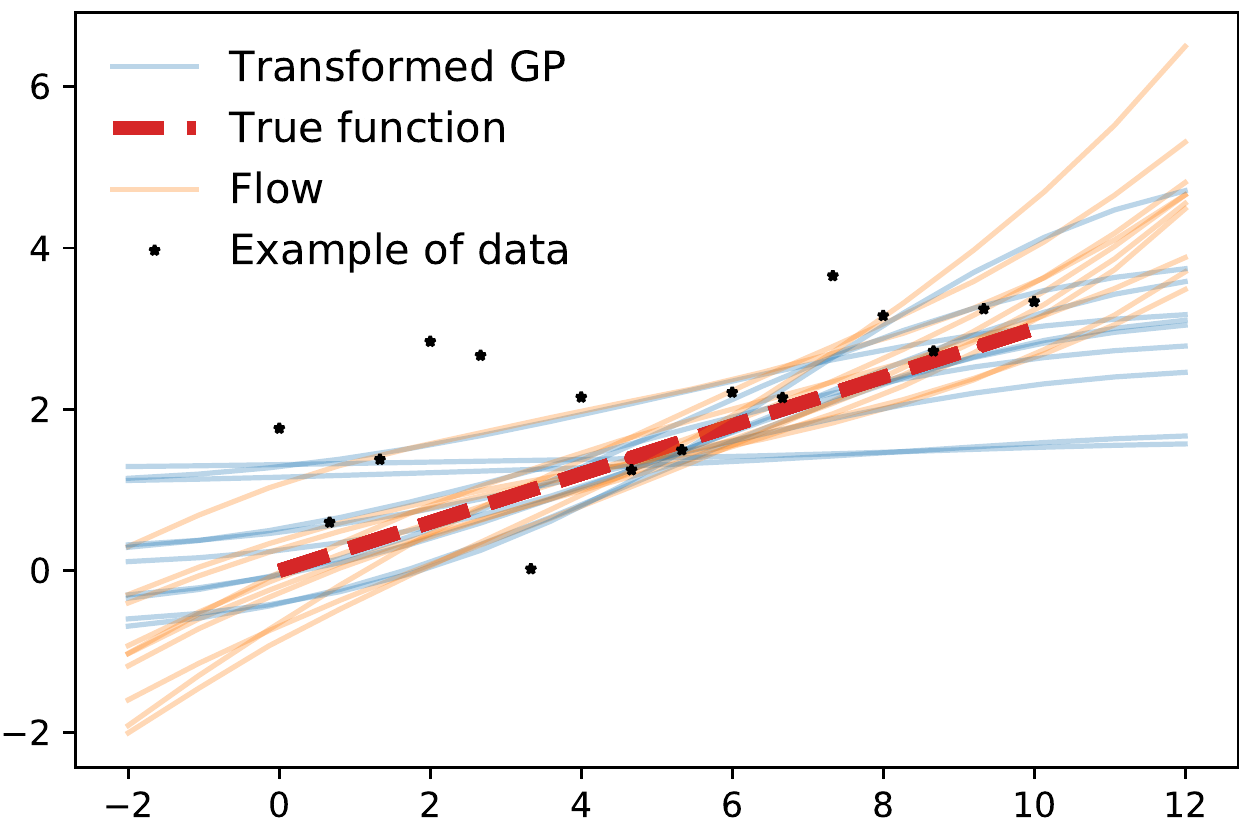}
        \caption{Linear, 15 data points}
    \end{subfigure} 
    \begin{subfigure}[h]{0.31\textwidth}
        \includegraphics[width=\textwidth]{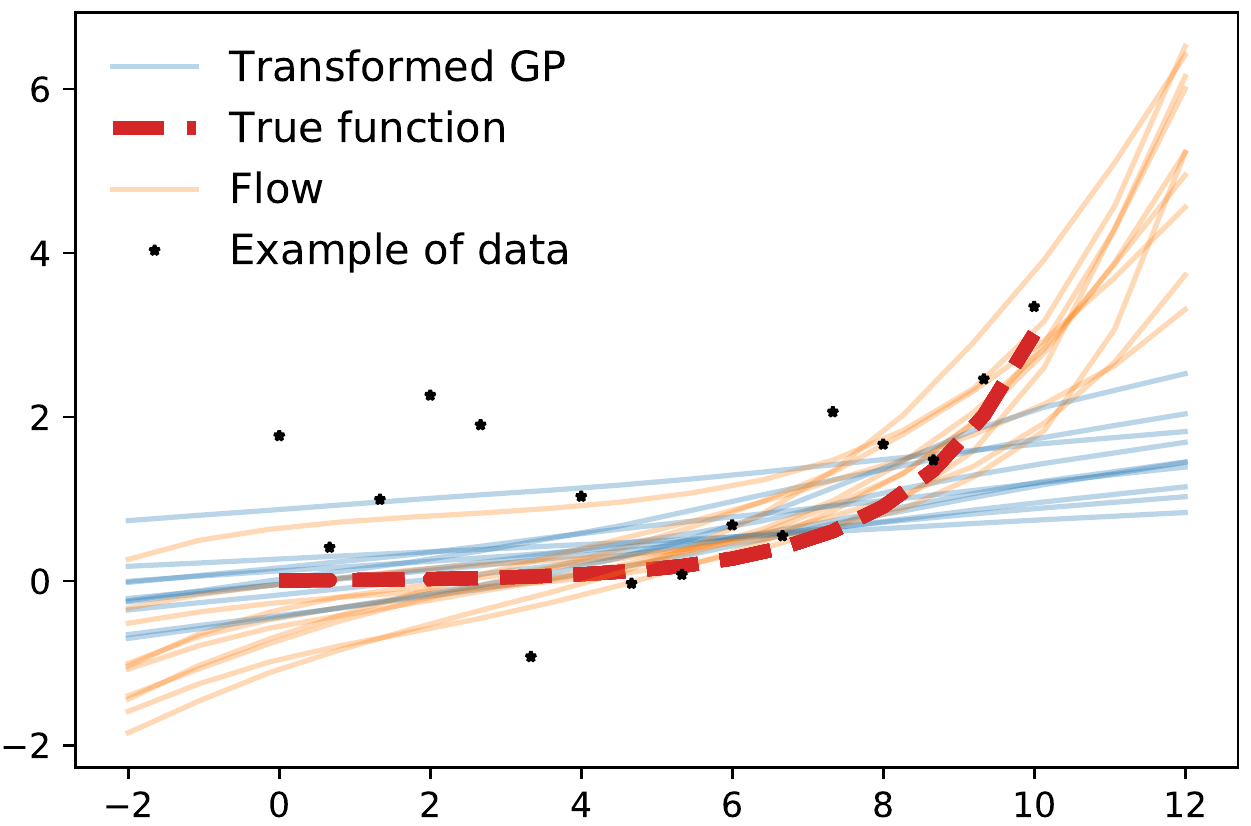}
        \caption{Exponential, 15 data points}
    \end{subfigure} 
    \begin{subfigure}[h]{0.31\textwidth}
        \includegraphics[width=\textwidth]{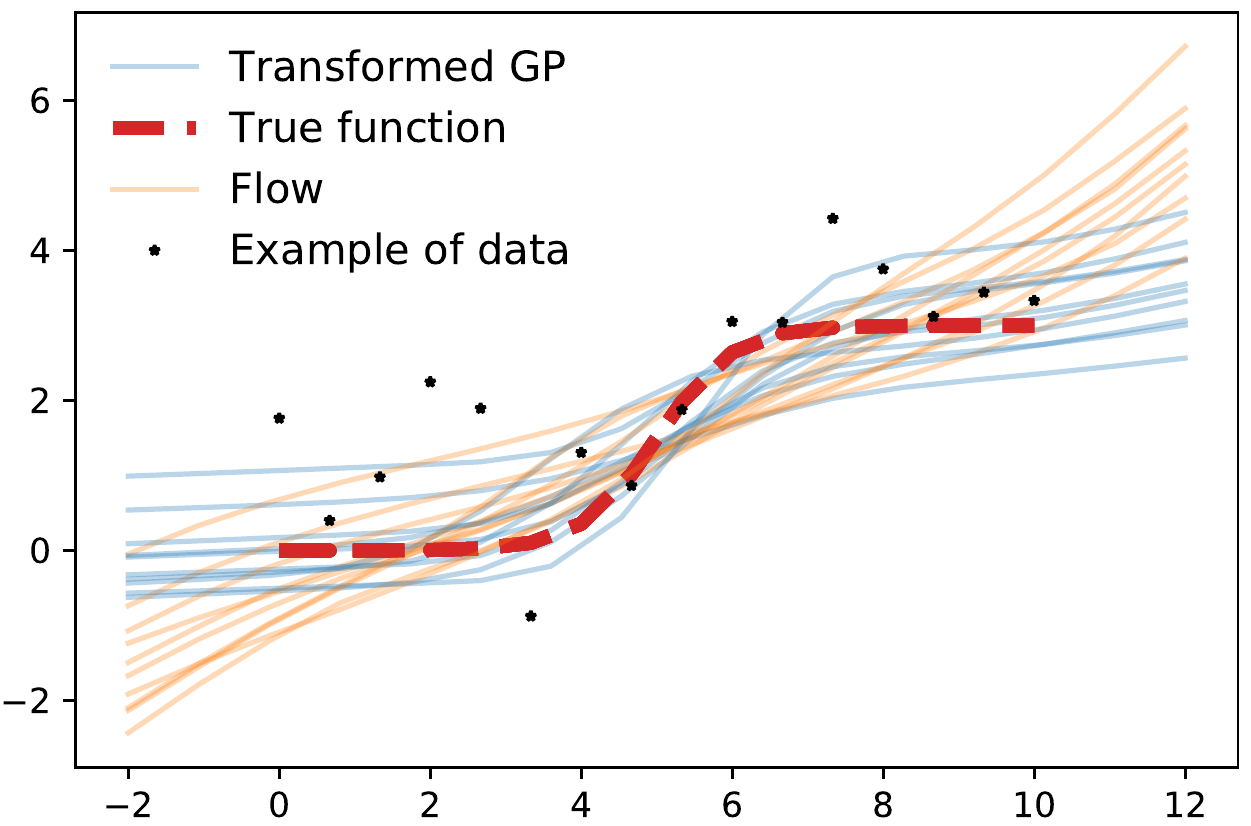}
        \caption{Logistic, 15 data points}
    \end{subfigure} \\
    \begin{subfigure}[h]{0.31\textwidth}
        \includegraphics[width=\textwidth]{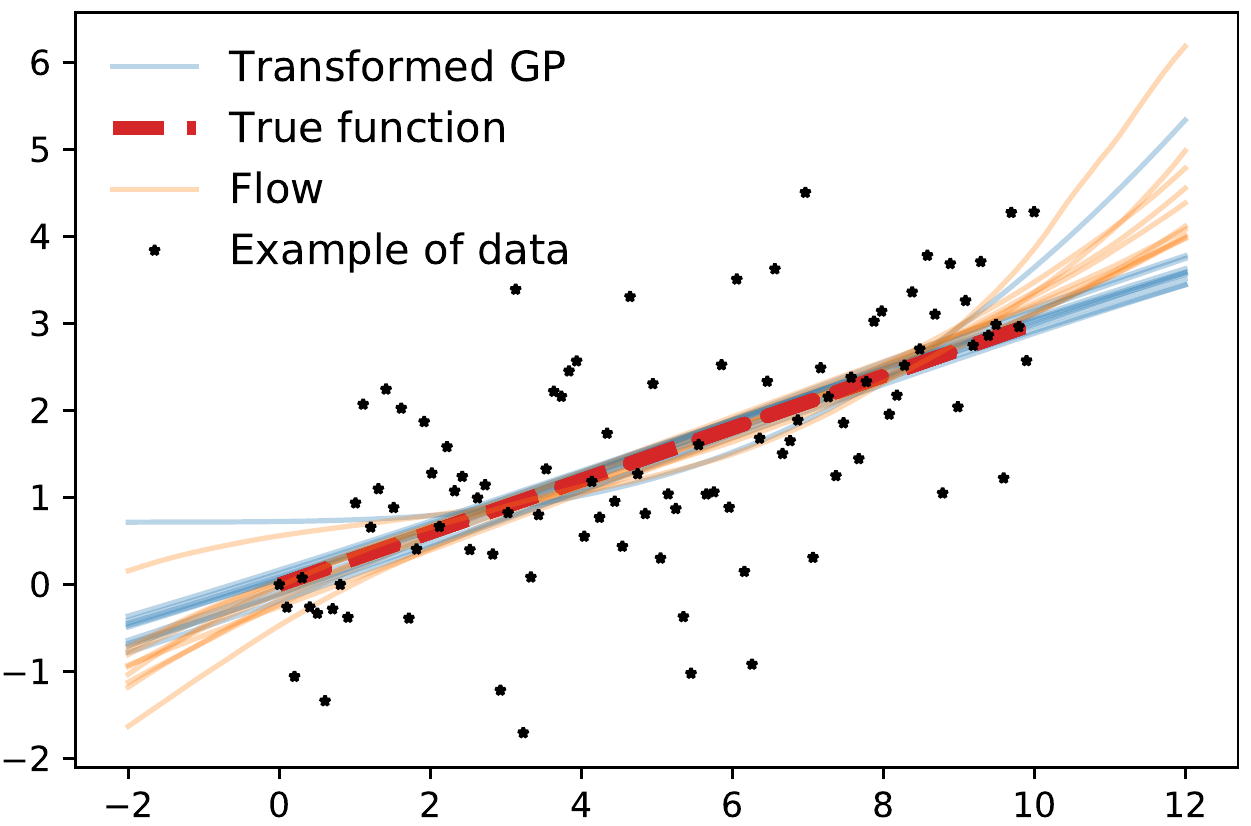}
        \caption{Linear, 100 data points}
    \end{subfigure} 
    \begin{subfigure}[h]{0.31\textwidth}
        \includegraphics[width=\textwidth]{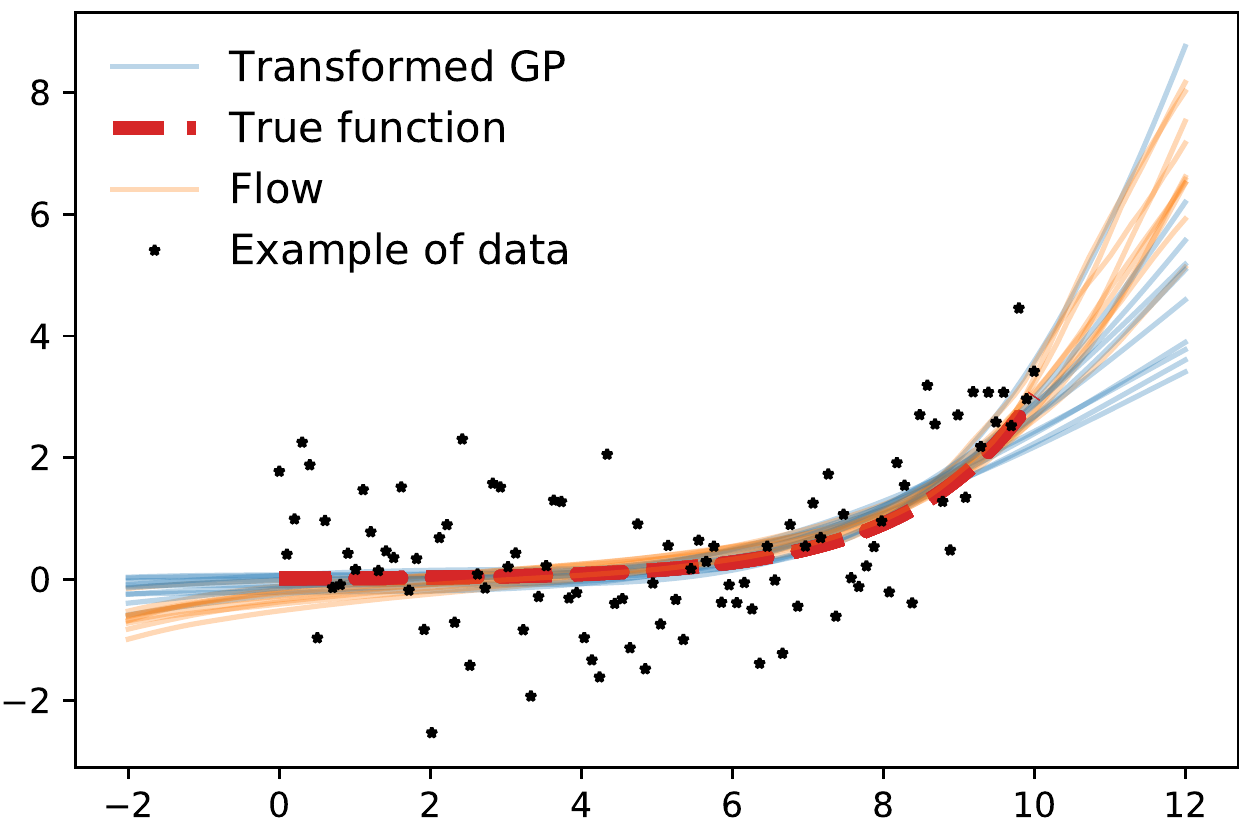}
        \caption{Exponential, 100 data points}
    \end{subfigure} 
    \begin{subfigure}[h]{0.31\textwidth}
        \includegraphics[width=\textwidth]{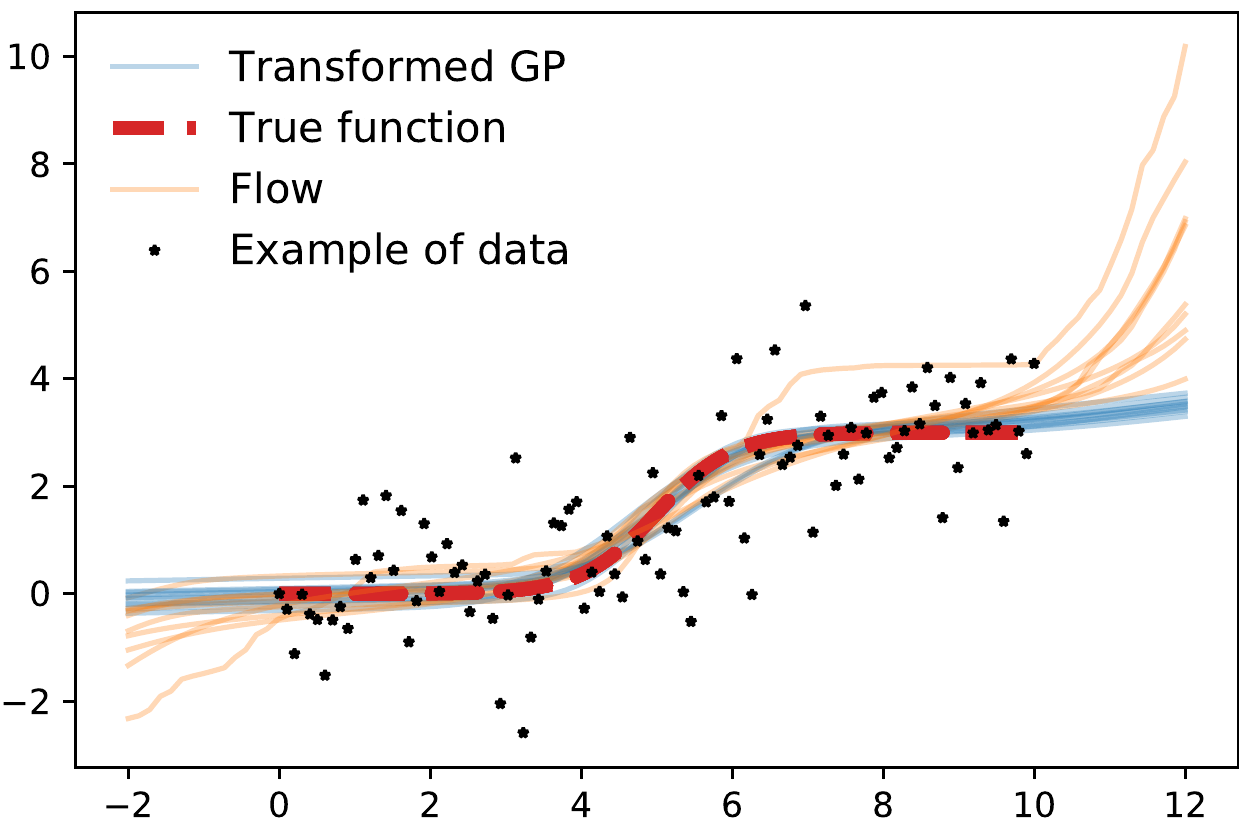}
        \caption{Logistic, 100 data points}
    \end{subfigure} 
    \caption{Mean fits for $10$ trials with different random noise as estimated by the flow and the transformed GP~\cite{Andersen:2018} (the noise samples are identical for both methods; we plot the data from one trial).}
    \label{fig:15_pts}    
\label{fig:monotonic-regression}
\end{figure*}

\paragraph{Uncertainty quantification in monotonic random processes}

In standard (non-monotonic) regression, GPs are used as the gold standard for the quantification of uncertainty~\cite{Foong:2019}. However, directly comparing the confidence intervals of a monotonic random process to a standard GP is misleading due to the additional constraints of monotonicity which lead to tighter confidence intervals as fewer explanations (functions) are compatible with the observed data. Fig.~\ref{fig:confidence-intervals} illustrates the shrinking of the confidence intervals for monotonic random processes in comparison to a standard (unconstrained) GP. 
As a baseline, we fit a standard GP (Fig.~\ref{fig:CI:1}) and consider only those samples from the posterior which are monotonic increasing in the domain in which we perform extrapolation ($[-5, 5]$); these samples along with their mean and 2 SD away from the mean are shown in Fig.~\ref{fig:CI:2}\footnote{We note that plotting the error bars using a Gaussian density may be misleading in monotonic regression as the samples from such process may not be symmetric around the mean, especially when the data are nearly constant, which can be seen by looking at the samples.}. % from these processes. One way to obtain more illustrative error bars in these situations is to use violin plots instead.}. 
The GP with monotonicity information %~\cite{Riihimaki:2010}
(Fig.~\ref{fig:CI:3}) is not able to guarantee that the samples are monotonic, especially in parts of the domain away from the data, while the transformed GP %~\cite{Andersen:2018}
(Fig.~\ref{fig:CI:4}) tends to underestimate the uncertainty, potentially due to the Dirichlet conditions imposed on the boundaries of the domain.
Meanwhile, the uncertainty estimates of our proposed monotonic flow are comparable to the baseline (i.e.~the monotonic samples from a standard GP) during extrapolation and samples from the flow are guaranteed monotone. %to be monotonic.

\begin{figure}[t]
\centering
    \begin{subfigure}[t]{0.46\textwidth}
        \includegraphics[width=\textwidth]{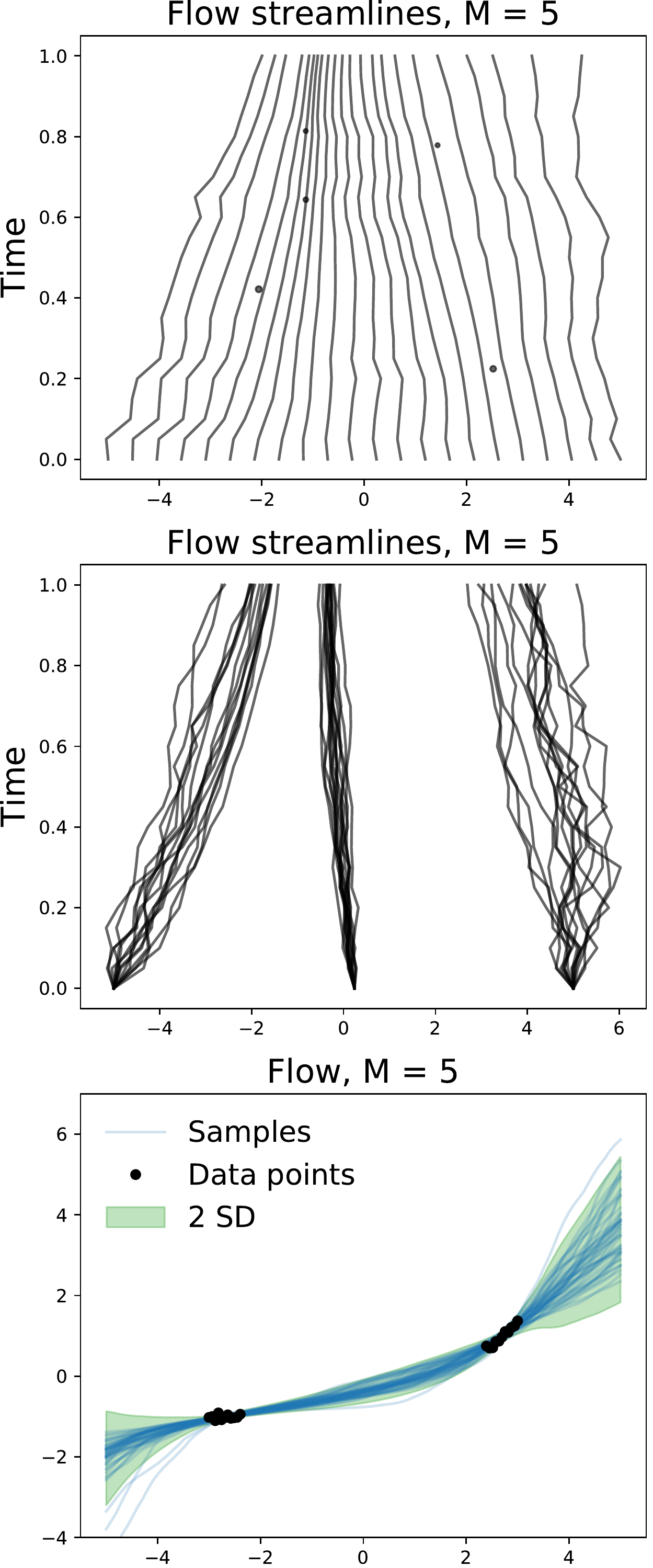}
        \caption{Flow streamlines for 5 inducing points.} \label{fig:streamlines:1}
    \end{subfigure} \hfill%\hspace{2pt}
    \begin{subfigure}[t]{0.46\textwidth}
        \includegraphics[width=\textwidth]{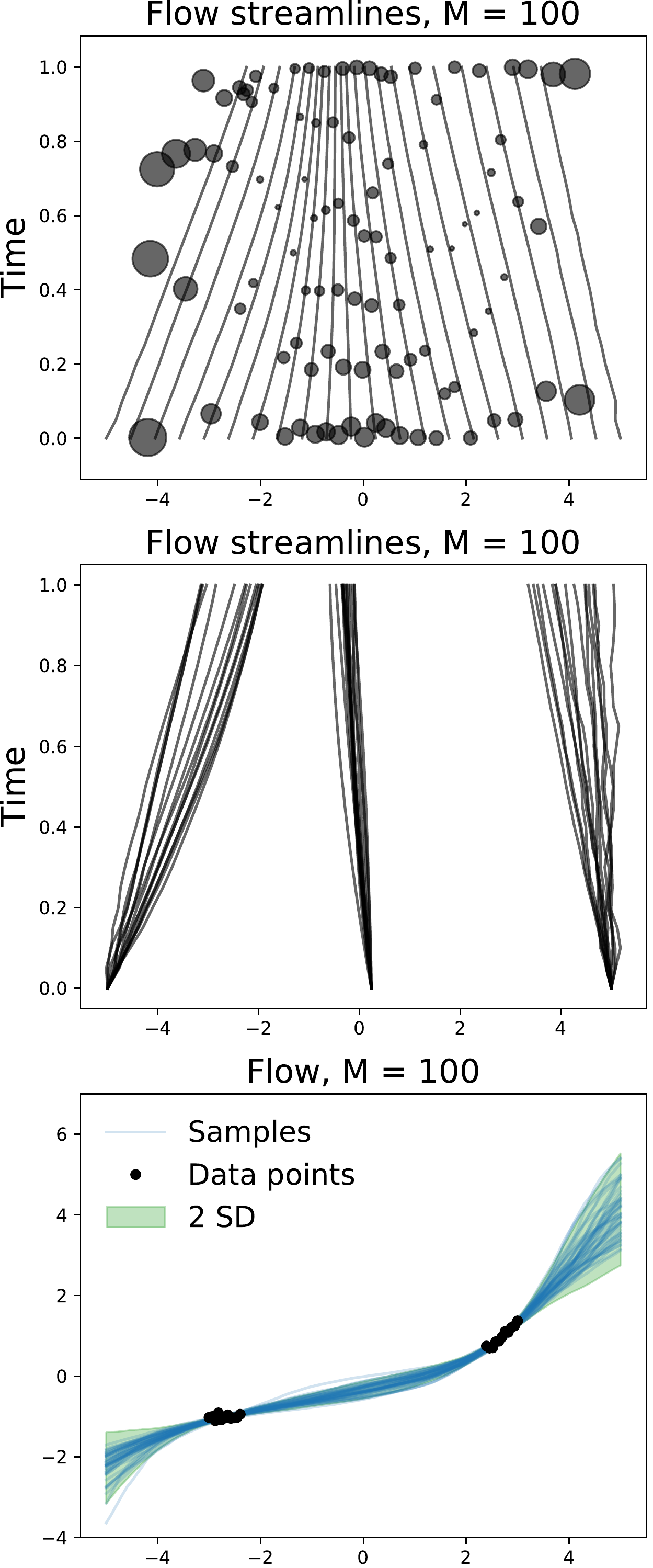}
        \caption{Flow streamlines for 100 inducing points.}
        \label{fig:streamlines:2}
    \end{subfigure}
    \caption{A coherent sample (top) and a set of independent samples at three chosen input locations (middle) from a fitted flow (bottom). The circles (top figures) show the location $\mm$ of the inducing points and are scaled by their (relative) variance $\Sb$.}
    \label{fig:streamlines}
\end{figure}        

\begin{figure}[t]
\centering
    \begin{subfigure}[t]{0.46\textwidth}
        \includegraphics[width=\textwidth]{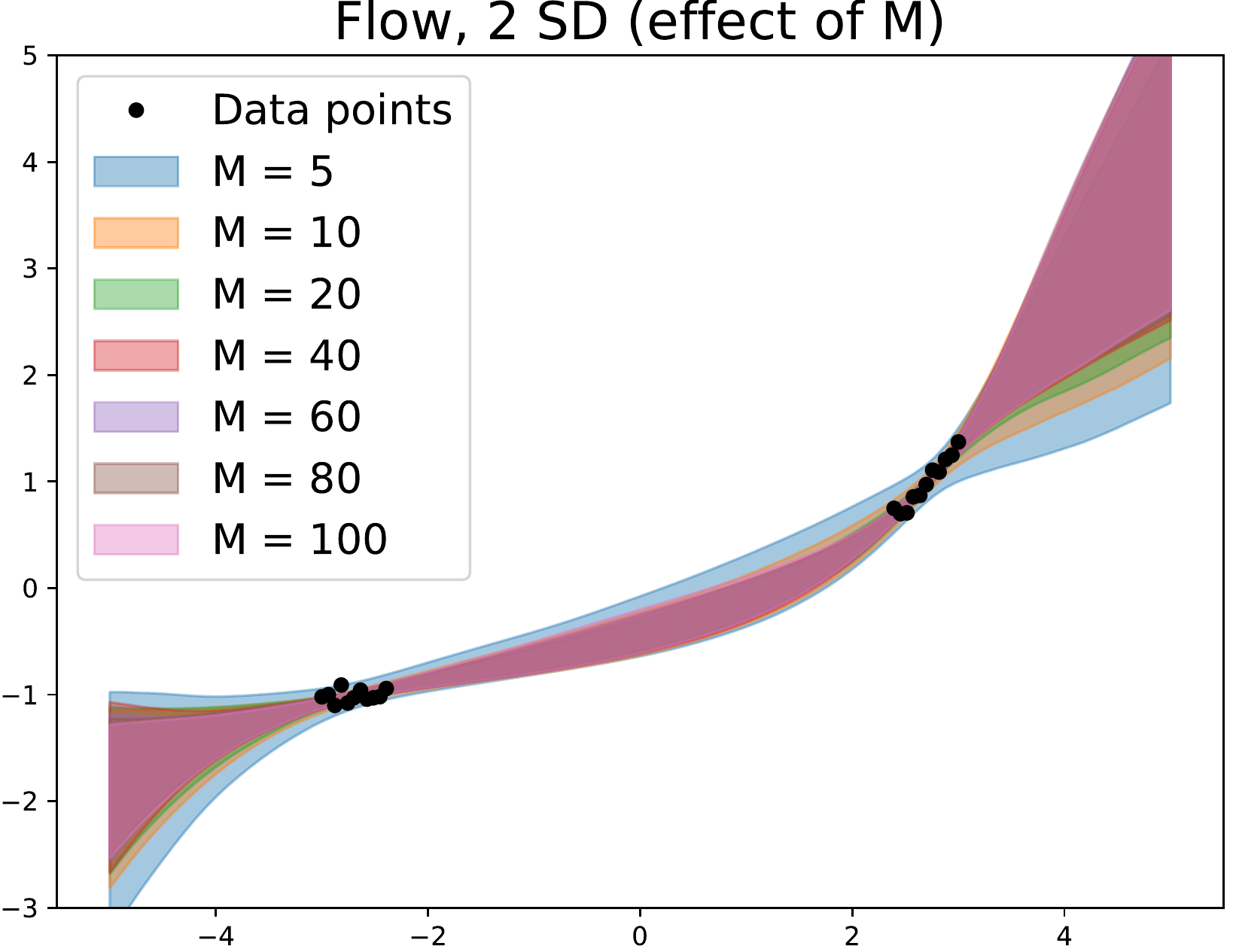}
        \caption{Flow comparison for $M = 5, 50, 100$.} \label{fig:streamlines:5}
    \end{subfigure} \quad
    \begin{subfigure}[t]{0.46\textwidth}
        \includegraphics[width=\textwidth]{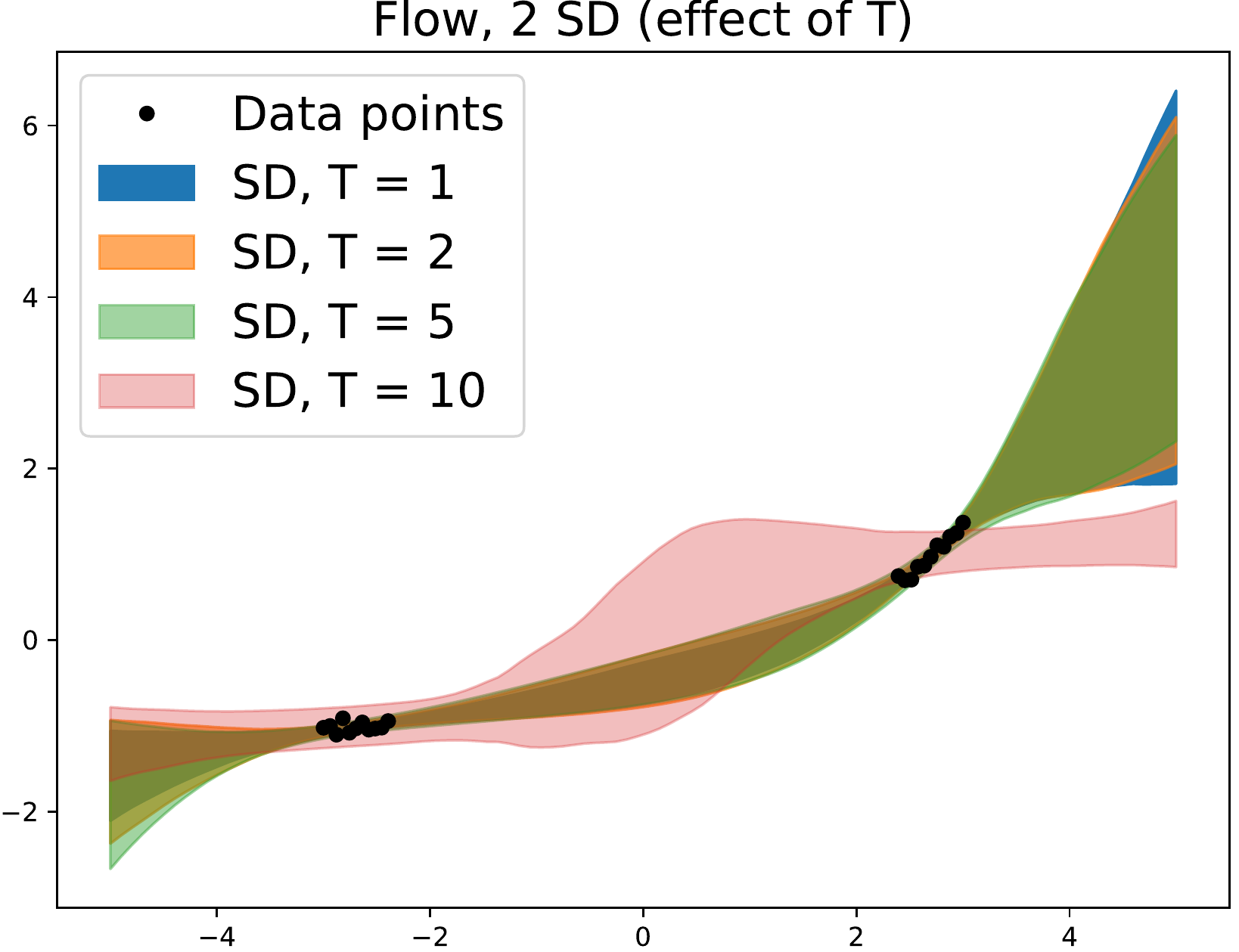}
        \caption{Flow comparison for $T = 1, 2, 5, 10$.} \label{fig:streamlines:6}
    \end{subfigure} 
    \caption{Effect of the number M of inducing points and the total flow time T on the estimated uncertainty (coloured regions correspond to 2 SD away from the mean of the samples from the flow). Results for 5 random trials.}
    \label{fig:flow_M_T}
\end{figure}
An alternative visualisation of the flow model involves looking at the streamlines of the input values as a function of time (see Fig.~\ref{fig:streamlines}). 
The streamlines may be visualised as one coherent draw from the flow (shown at the top of Fig.~\ref{fig:streamlines}), or as independent samples at a given value of the inputs (shown in the middle figures of Fig.~\ref{fig:streamlines}).
The latter also help visualise the uncertainty in the model as these samples show the range of possible outputs $S(T, \omega; \x)$ for a given input location $\x$. 

The mean and variance of the inducing points in the flow GP depend on their number $M$ as follows: given few inducing points, they are typically optimised to be located close to the observations so that the resulting model fits the observations well (with low estimated observational noise).
Meanwhile, given a large number of inducing points, some are used to fit the data well while others are placed in regions with no observations (see, for example, the regions in between the data ($[-2, 2]$) in Fig.~\ref{fig:streamlines:2}) and optimised to have higher variance $\Sb$ in those regions. 
We note that the uncertainty estimates in the monotonic flow model do not depend much on the number of inducing points: the estimates are nearly identical for $M > 5$ while for this data $M = 5$ may not be enough to explain the data well, hence the observational noise gets overestimated, also resulting in higher variance in extrapolation.
Fig.~\ref{fig:streamlines:5} shows how the uncertainty estimates for this data set depend on the number of inducing points. Similarly, Fig.~\ref{fig:streamlines:6} details the dependence on the flow time $T$~\cite{Hedge:2019}; longer flow time ($T \geq 10$) results in more extreme warpings and thus higher uncertainty at the observations with overestimated observational noise.

\section{ALIGNMENT APPLICATION} 
\begin{figure}[t]
    \includegraphics[width=0.9\linewidth]{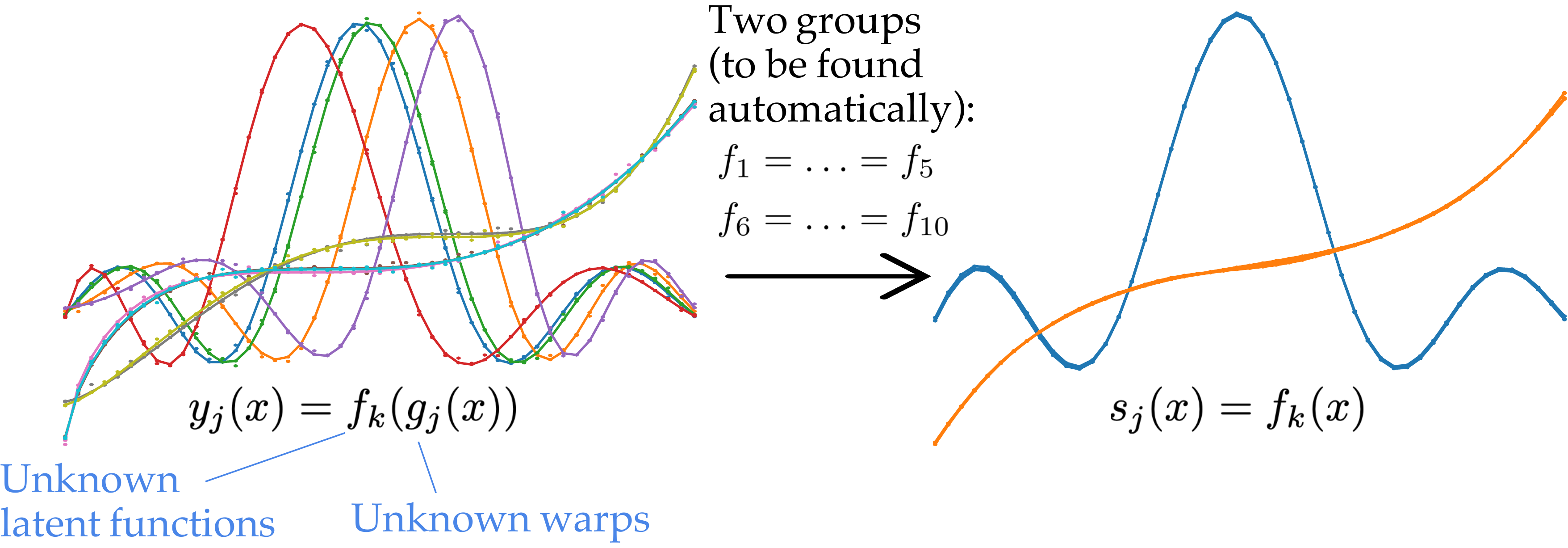}
    \caption{Illustration of the alignment problem.}
\end{figure} \label{fig:alignment_toy}
\begin{figure}[h!]
\centering 
    \begin{subfigure}[h]{.3\textwidth}
            \includegraphics[width=\textwidth]{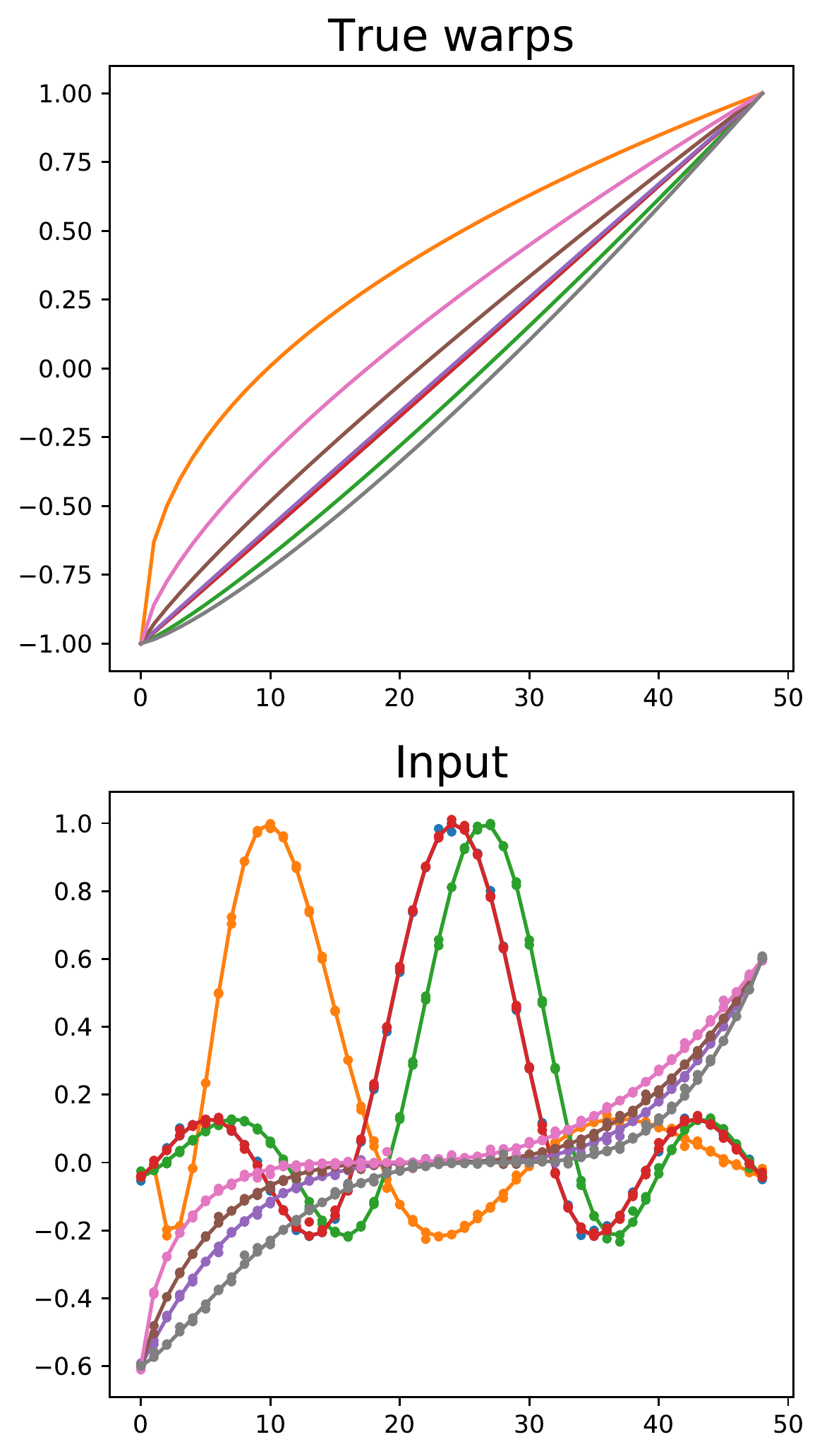}
    \end{subfigure} \vrule
    \begin{subfigure}[h]{0.32\textwidth}
            \includegraphics[width=\textwidth]{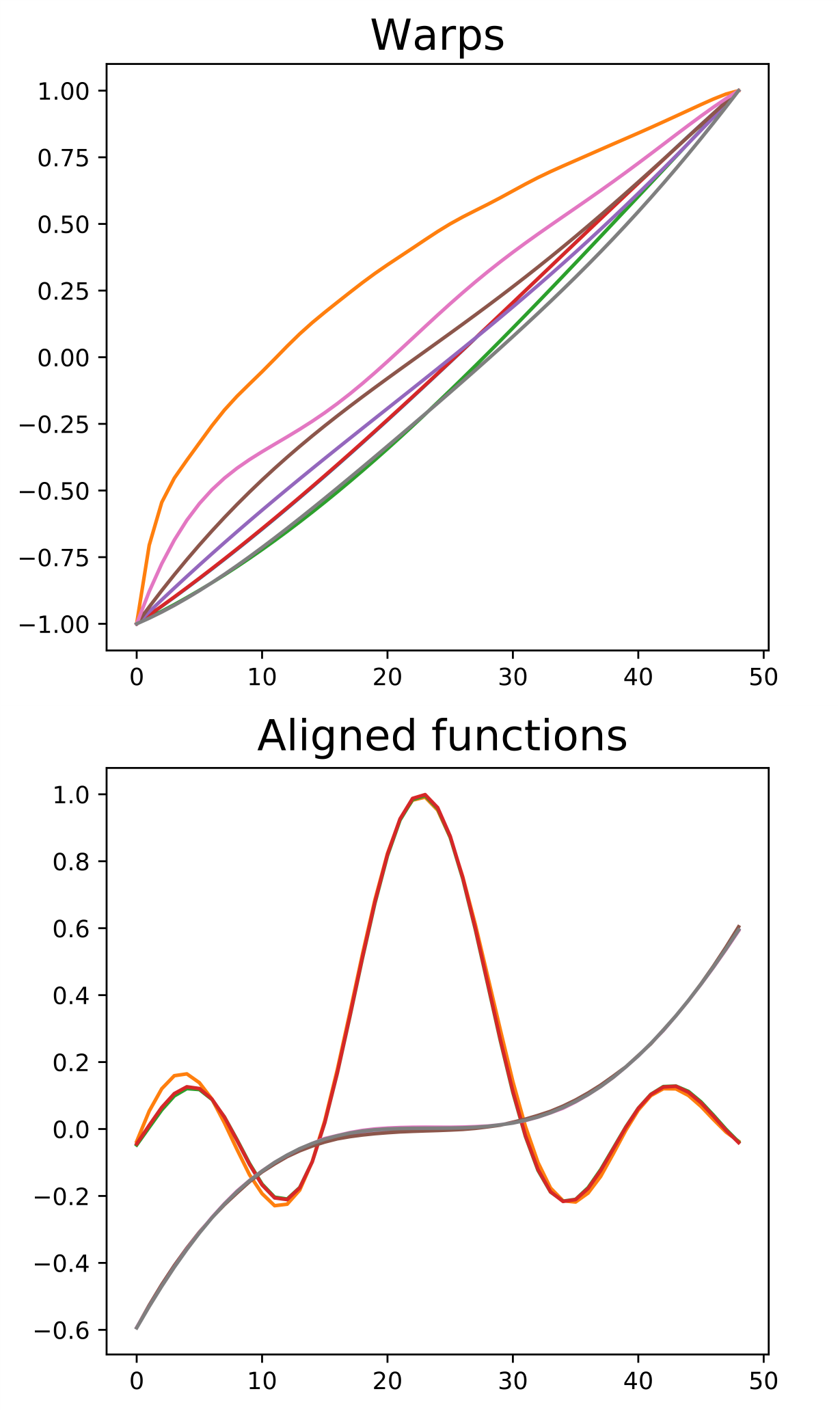}
    \end{subfigure}
    \begin{subfigure}[h]{0.3\textwidth}
            \includegraphics[width=\textwidth]{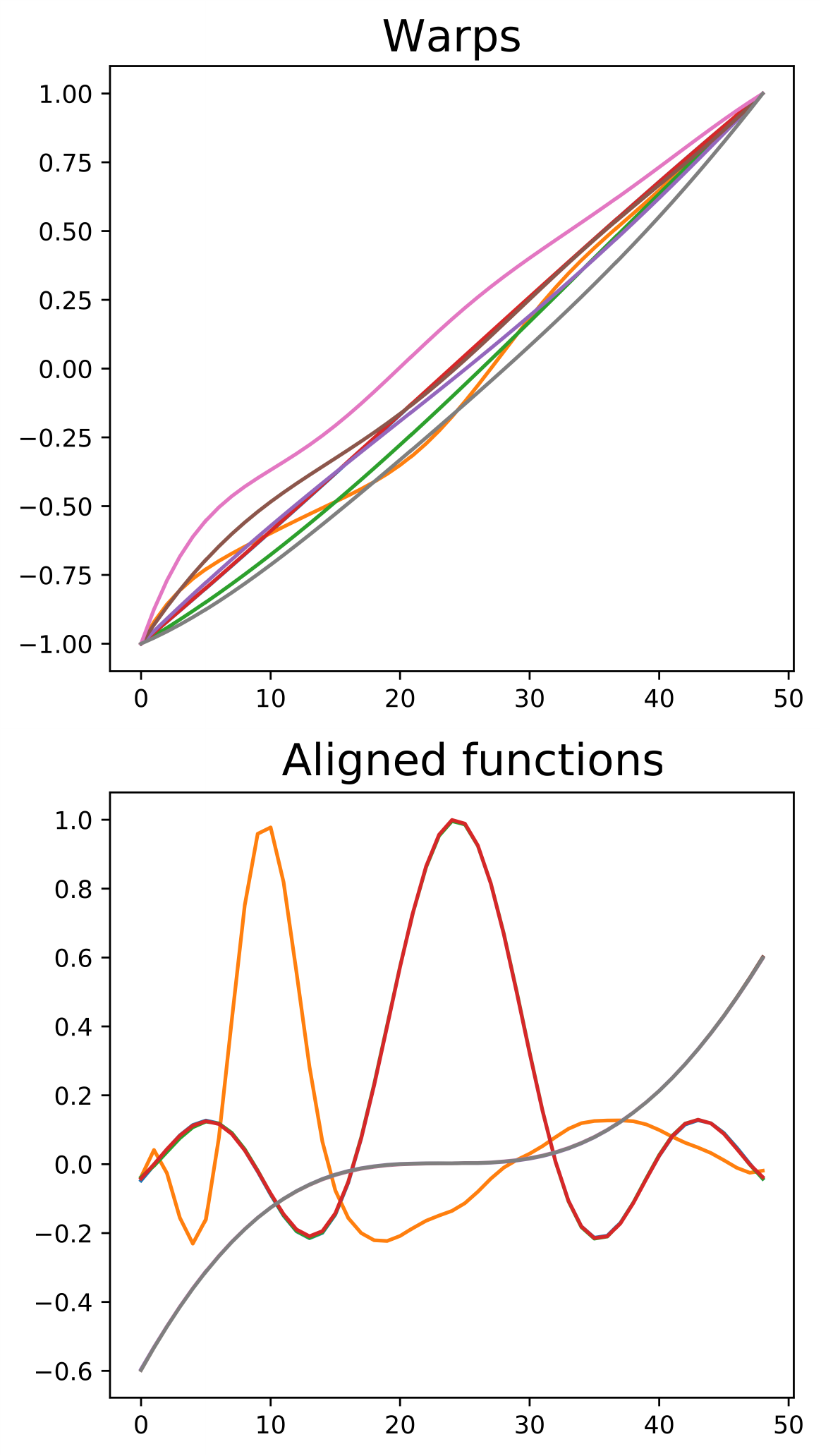}
    \end{subfigure}
    \caption{The observations (bottom left) were generated by applying warping functions (top left) to a sinc or cubic function ($K = 2$). The point estimates in the alignment model of~\cite{Kazlauskaite:2019} will return one of the two possible solutions (middle and right columns). One solution (middle) aligns all the sequences into two groups but uses more extreme warps (note the orange warp) while the other solution (right) assigns the orange sequence to a new cluster (and thus uses an identity warp for this sequence). Both are plausible given our priors on the warps $g_j(\cdot)$ and the functions $f_j(\cdot)$, hence a preferred model would preserve the uncertainty about the warps and the cluster assignments and capture the full range of possible solutions.} 
    \label{fig:uncertainty}
\end{figure}

\begin{figure*}[h!]
    \centering
    \floatbox[{\capbeside\thisfloatsetup{capbesideposition={left,bottom},capbesidewidth=4cm}}]{figure}[\FBwidth]
    {\caption{Given noisy observations of warped sequences, we compare the uncertainty in the warps for the proposed model (with and without correlations) and for~\cite{Kazlauskaite:2019}.}\label{fig:uncert:align}}
    {\hspace{-4.6cm}\includegraphics[scale=0.25]{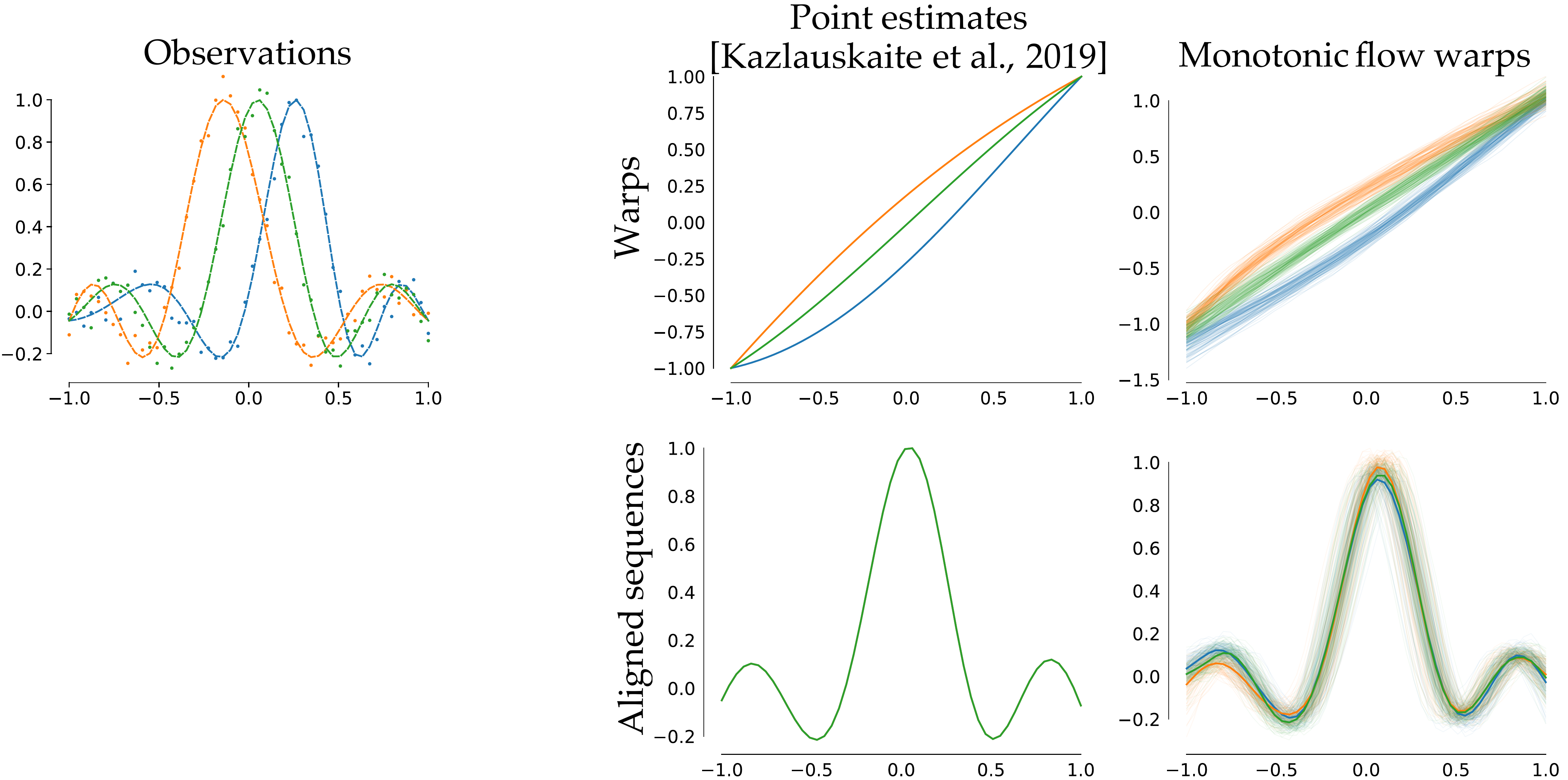}}
\end{figure*}

A monotonic constraint in the first layer is desirable in mixed effects models where the first layer corresponds to a warping of space or time that does not allow permutations. We consider an application of the monotonic random process as an integral part of a model designed to align multiple temporal sequences of observations.
This problem is introduced in detail in~\cite{Kazlauskaite:2019} and here we provide a short summary and the baseline model.
We change notation 
to match~\cite{Kazlauskaite:2019}; for the alignment application, $g(\cdot)$ refers to the monotonic function, specified using the monotonic Gaussian process flow model, whereas $f(\cdot)$ is now an arbitrary function.

\newcommand{\f}{\mathbf{f}}
\newcommand{\g}{\mathbf{g}}

Assume we are given some time-series data with inputs $\x \in \mathbb{R}^N$ and $J$ output sequences $\{\y_j \in \mathbb{R}^N\}_{j=1}^{J}$. 
We know that there are multiple underlying functions that generated this data, say $K$ such functions, $f_k(\cdot)$, and the observed data were generated by warping the temporal inputs to the true functions using a monotonic warping function $g_j(\x)$, such that:
\begin{gather}
    \y_j = f_k\big( g_j(\x) \big) + \epsilon_j.  
\end{gather}
where $\epsilon_j \sim \NN(0, \beta^{-1} I_N )$ is observation noise. 
Then the corresponding, latent, sequences that are not corrupted by the temporal warp (\emph{i.e.}~the aligned versions of $\y_j$) are $\f_j := f_k(\x)$. The functions $f_j(\cdot)$ are modelled jointly using a GP and the joint conditional likelihood for each pair of sequences, $\y_j$ and $\f_j$, is:
\begin{equation}
\begin{aligned}
p & \left(\begin{bmatrix} \f_j\\ \mathbf{y}_j\end{bmatrix}  \middle| \, \g_j, \x, \theta_j \right)  \\ 
&\sim 
\mathcal{N}\left(\mathbf{0}, 
\begin{bmatrix} k_{\theta_j}(\x, \x) & k_{\theta_j}(\x, \g_j) \\ k_{\theta_j}(\g_j, \x) & k_{\theta_j}(\g_j, \g_j) + \beta_j^{-1}  \end{bmatrix} 
        \right) %\nonumber 
        \label{eq:GPGPLVM:model_over_time}
\end{aligned}
\end{equation} where $\g_j := g_j(\x)$ are finite-dimensional realisations of the warping function, and $\theta_j$ includes the parameters of the GP that models function $f_j(\cdot)$ and the parameters of the warping function $g_j(\cdot)$. The task is then to learn the latent functions $f_k(\cdot)$ and the warps $g_j(\cdot)$ such that the versions of these function which are not corrupted by the warp, $\f_j$, are aligned as well as possible. Note that the number $K$ of distinct functions $f_k(\cdot)$ is unknown must also be inferred from the data. 
This is achieved by formulating an alignment objective that pushes the uncorrupted sequences $\f_j$ into $K$ groups where each group corresponds to one latent function $f_k(\cdot)$, and the sequences within each group are aligned to each other. If the warps fully explain the differences between the sequences within each group, then each group contains a single sequence $\f_k(\cdot)$ (or, equivalently, all the sequences within a group coincide); see Fig.~\ref{fig:alignment_toy}.

\newcommand{\F}{\mathbf{F}}
\renewcommand{\v}{\mathbf{v}}

Previously, \cite{Kazlauskaite:2019}~proposed a probabilistic alignment objective based on a GP latent variable model (GP-LVM)~\cite{Lawrence:2004} that aligns sequences within groups. A GP-LVM is a generative model that is often used as a dimensionality reduction technique to uncovers the latent structure in the data by constructing a low dimensional manifold, and using independent GPs as mappings from a latent space to an observed space. 
In a GP-LVM, GPs are taken to be independent across the features (columns) of the data $\F = \{\f_j\}_{j=1}^{J} \in \mathbb{R}^{J \times N}$, and the likelihood function is:
\begin{equation}
    p(\F \given \v) %= \prod_{n=1}^N p(\F_{:,n} \given \v) 
    = \textstyle\prod_{n=1}^N \mathcal{N}(\F_{:,n} \given 0, k(\v, \v) + \hat{\beta}^{-1} I_{J})\label{eq:GPLVM_model}
\end{equation}
where $\v = \{v_j\}_{j=1}^{J}$ are the latent variables for each sequence, the GP covariance is $k(\v, \v)$ and $\hat{\beta}$ is a noise precision. 
Typically, a GP-LVM contains a prior on the latent variables $p(\v) \sim \mathcal{N}(\mathbf{0}, \sigma^{2}_v I_J)$. 
This encourages the latent variables to be placed close to the origin in the latent space if the observed data can be explained by a single cluster in a latent space; otherwise a minimal number of clusters of $\v$ in the latent space will be made under the Bayesian prior encouraging sparsity. 
Furthermore, the stationary kernel in the GPs that map from the latent space to the observed space depend only on the distance between the latent location $v_j$, which means that the further each latent location is from the others, the less correlated the corresponding GP outputs $\f_j$. Specifically, this behaviour is controlled by the kernel length-scale which allows to reduce correlation between groups of sequences while maintaining strong correlation among sequences within each group. 

The
two objectives of (\ref{eq:GPGPLVM:model_over_time}) and (\ref{eq:GPLVM_model}) 
are combined
to learn the GPs for $f_j(\cdot)$, the warps $g_j(\cdot)$ and the number of underlying clusters $K$. The baseline~\cite{Kazlauskaite:2019} uses MAP estimates for the fitting of the GPs in (\ref{eq:GPGPLVM:model_over_time}) and for the GP-LVM in (\ref{eq:GPLVM_model}). This leads to point estimates of the warps $\g_j$ and, consequently, does not retain any information about the uncertainty of cluster assignments. Fig.~\ref{fig:uncertainty} illustrates the limitation of using point estimates; the observed data can be explained in multiple different ways which cannot be uncovered using point estimates. 

\paragraph{Uncertainty in monotonic warps} We demonstrate the ability of our monotonic random process to capture the uncertainties in the warps and the cluster assignments in the alignment model. In order to preserve the compositional uncertainty in both the warping functions $g_j(\cdot)$ and the latent functions $f_j(\cdot)$, we introduce correlations between the variational distributions in the two layers, $g(\cdot)$ and $f(\cdot)$ using the inference scheme detailed in~\cite{Ustyuzhaninov:2019}. 
Fig.~\ref{fig:uncert:align} illustrates this phenomenon, and compares the uncertainty in the warps for the original point estimate~\cite{Kazlauskaite:2019}, the monotonic flow and the flow with correlations between the samples from the warp and the function $f$. The flow captures a range of different possible warpings that are consistent with our prior (which favours solutions that are close to an identity warp) and also fits and aligns the data well. An additional example with bi-modal behaviour, as in Fig.~\ref{fig:uncertainty}, is given in Fig.~\ref{fig:supp:align} in the Supplement.

% !TEX root = ../aistats2020.tex
%\section{Conclusion} 
%\label{sec:conclusions}

\section{CONCLUSION}
We have proposed a novel \nonparametric{} model of monotonic functions based on a random process with monotonic trajectories that confers improved performance over the state-of-the-art as well as preferable theoretical properties. 
Many real-life regression tasks deal with functions that are known to be monotonic, and explicitly imposing this constraint helps uncover the structure in the data, especially when the observations are noisy or data are scarce.
We have also demonstrated that the proposed construction can be used as part of a complex alignment model where the uncertainty estimates provide a more informative model and help uncover structures in the data that are not captured by the existing models.
% We have also demonstrated that monotonicity constraints can be used to guard against degenerate solutions and improve the interpretability of multi-layer GP-based models.
% %
More broadly, with additional mid-hierarchy marginal information or domain specific knowledge of compositional priors, %~\cite{Kaiser:2018, Kazlauskaite:2019}
\emph{e.g.}~\cite{Kaiser:2018}, hierarchical models may necessitate a composition of (injective) monotonic mappings for all but the output layer. 
This advocates the study of %nonparametric models of 
monotonic functions, which can represent a wide variety of transformations and hence serve as a general purpose first layer in a hierarchical model, especially, when the function is known to be non-stationary.

% Monotonicity also appears in the more general context of hierarchical models where we want to transform a (simple and typically stationary) input distribution to a (complicated and non-stationary) data distribution. This may necessitate a composition of injective mappings for all but the output layer; this property is met if the hidden layers in the model comprise monotonic transformations. 
% In the presence of additional mid-hierarchy marginal information or domain specific knowledge of compositional priors, e.g.~\cite{Kaiser:2018, Kazlauskaite:2019}, the hierarchy allows us to exploit our prior beliefs.
% In the absence of such information (e.g.~unsupervised learning), multiple injective layers would only arise if the individual layers were insufficiently expressive to capture the transformation (as the concatenation of injective transforms is itself injective).
% That motivates the study of nonparametric models of monotonic (injective) functions, which can represent a wide variety of transformations and hence serve as a general purpose first layer in a hierarchical model.
\newpage
\newpage

\subsubsection*{Acknowledgments}
This work has been supported by EPSRC CDE (EP/L016540/1) and CAMERA (EP/M023281/1) grants as well as the Royal Society. The authors are grateful to Markus Kaiser and Garoe Dorta for their insight and feedback on this work, and to Michael Andersen for sharing the implementation of Transformed GPs. IK would like to thank the Frostbite Physics team at EA.

% \FloatBarrier

% \balance

\bibliography{references}
\bibliographystyle{apalike} % {plain} % Guidelines want a full author name + year
\newpage
\renewcommand{\thesection}{A}

\setcounter{figure}{0}
\renewcommand\thefigure{\thesection\arabic{figure}}

\section*{Supplementary material}

\renewcommand\floatpagefraction{.9}
\renewcommand\dblfloatpagefraction{.9} % for two column documents
\renewcommand\topfraction{.9}
\renewcommand\dbltopfraction{.9} % for two column documents
\renewcommand\bottomfraction{.9}
\renewcommand\textfraction{.1}   
\setcounter{totalnumber}{50}
\setcounter{topnumber}{50}
\setcounter{bottomnumber}{50}

% \section*{Monotonic Gaussian Process Flows: Supplementary material}

\subsection{Numerical solution of the SDE} \label{sec:numerical_solution}
To ensure that the SDE solutions are monotonic functions of the initial values, we make assumptions about the Wiener process realisations $W(\cdot, \omega)$. 
To compute the SDE solutions under such assumptions, we draw a Wiener process realisation as well as the flow field drift and diffusion, and given these draws, we use the Euler-Maryama numerical solver (following~\cite{Hedge:2019}).
Specifically, starting with the initial state $(x = x_1, t = 0), \ldots, (x = x_N, t = 0)$, we use~\eqref{eq:gp-posterior} to compute the drift and diffusion at the current state, and the discretised version of~\eqref{eq:sde} (\emph{i.e.}~with $\Delta t$ and $\Delta W$ instead of $dt$ and $dW$) to compute the state update $\Delta x$. 
This gives the new state $(x_1 + \Delta x_1, \Delta t),...,(x_n + \Delta x_n, \Delta t)$, and repeating this procedure $(T / \Delta t)$ times, we arrive at the state $(S(T, \omega; x_1), T),...,(S(T, \omega; x_N), T)$, corresponding to the approximate SDE solution at time T. 
The monotonic trajectories are recovered by the numerical SDE solver in the limit of the step size going to zero, $\Delta t \rightarrow 0$. 
Therefore, the step size must be sufficiently small w.r.t. the smoothness of the flow; since we use a GP to define the flow, the smoothness is determined by the lengthscale of the kernel. 

\subsection{Implementation details}
Our model is implemented in Tensorflow~\cite{Tensorflow}.
For the evaluations in Tables~1 and 2 we use $10 000$ iterations with the learning rate of $0.01$ that gets reduced by a factor of $\sqrt{10}$ when the objective does not improve for more than $500$ iterations. For numerical solutions of SDE, we use Euler-Maruyama solver with $20$ time steps, as proposed in~\cite{Hedge:2019}. 

\subsection{Computational complexity} Computational complexity of drawing a sample from the monotonic flow model is $\mathcal{O}\big(N_\text{steps} (N M^2 + N^2)\big)$, where $N_\text{steps}$ is the number of steps in numerical computation of the approximate SDE solution, $N M^2$ is the complexity of computing the GP posterior for $N$ inputs based on $M$ inducing points, and $N^2$ is the complexity of drawing a sample from this posterior. We typically draw fewer than $5$ samples to limit the computational cost.

\subsection{Non-Gaussian noise} The inference procedures for the monotonic flow and for the 2-layer model can be easily applied to arbitrary likelihoods, because they are based on stochastic variational inference and do not require the closed form integrals of the likelihood density.

\subsection{Functions for evaluating the monotonic flow model}
\label{supp:regression_functions}
The functions we use for evaluations are the following:
\begin{itemize}[label={}]%,font=\small]%,
    \item $f_1(x) = 3, \; x \in (0, 10] $ \hspace{7pt} (flat function)
    \item $f_2(x) = 0.32 \: (x + \sin(x)),\; x \in (0, 10] $ \hspace{7pt} (sinusoidal function)
    \item $f_3(x) = 3 \; \text{if} \; x \in (0, 8], \; f_3(x) = 6 \; \text{if} \; x \in (8, 10] $ \hspace{7pt} (step function)
    \item $f_4(x) = 0.3 x, \; x \in (0, 10] $ \hspace{7pt} (linear function)
    \item $f_5(x) = 0.15 \: \exp(0.6x - 3), \; x \in (0, 10] $ \hspace{7pt} (exponential function)
    \item $f_6(x) = 3 \: / \: [1 + \exp(-2x + 10)] , \; x \in (0, 10] $ \hspace{7pt} (logistic function)
\end{itemize}
For the experiments shown in Fig.~3 we generate $50$ data points using $y = \text{sinc}(\pi x) + \varepsilon, \varepsilon \sim \NN(0, 0.02)$ for linearly spaced inputs $x \in [-1, 1]$.

\subsection{Regression evaluation parameters}
\label{supp:regression-parameters}

For the GP with monotonicity information we choose $M$ virtual points and place them equidistantly in the range of the data; we provide the best RMSEs for $M \in [10, 20, 50, 100]$. For the transformed GP we report the best results for the boundary conditions $L \in [10, 15, 20, 30]$ and the number of terms in the approximation $J \in [2, 3, 5, 10, 15, 20, 25, 30]$. For both models we use a squared exponential kernel. Our method depends on the time $T$, the kernel and the number of inducing points $M$. For this experiment, we consider $T \in [1, 5]$, $M = 40$ and two kernel options, squared exponential and ARD Mat{\'e}rn $3/2$. The lowest RMSE are achieved using the flow and the transformed GP.

\subsection{Uncertainty in alignment model}
\label{sec:alignment_uncertainty_supp}

To further illustrate the advantages of capturing the uncertainty about the warpings, we wish to find the possibly bi-modal warpings for each sequence. We use a Gaussian mixture model (instead of a single Gaussian) as the distribution of both, the warpings and the latent variables $\mathbf{Z}$ in the GP-LVM. In particular, the inducing points of the flow for each sequence are defined to be distributed as a mixture of two multivariate Gaussians. Then, given a draw from the Categorical distribution of this mixture, we defines the clusters assignments for each sample, and assign the resulting aligned sequences $\s_j$ to one of the coherent mixture component in the distribution of the latent points of the GP-LVM. Fig.~\ref{fig:supp:align} illustrates this behaviour, and gives an example where the uncertainty in the warps results in ambiguity in cluster assignments. 
A full discussion of the importance of correlations in the variational parameters for compositional uncertainty is available in~\cite{Ustyuzhaninov:2019} which provides further details of the inference scheme used.

\newdimen\figrasterwd
\figrasterwd\textwidth

\begin{figure*}[]
  \centering
  \parbox{\figrasterwd}{
    \parbox{.37\figrasterwd}{\centering%
      \subcaptionbox{Observations of 3 warped sequences.}{\includegraphics[width=0.8\hsize]{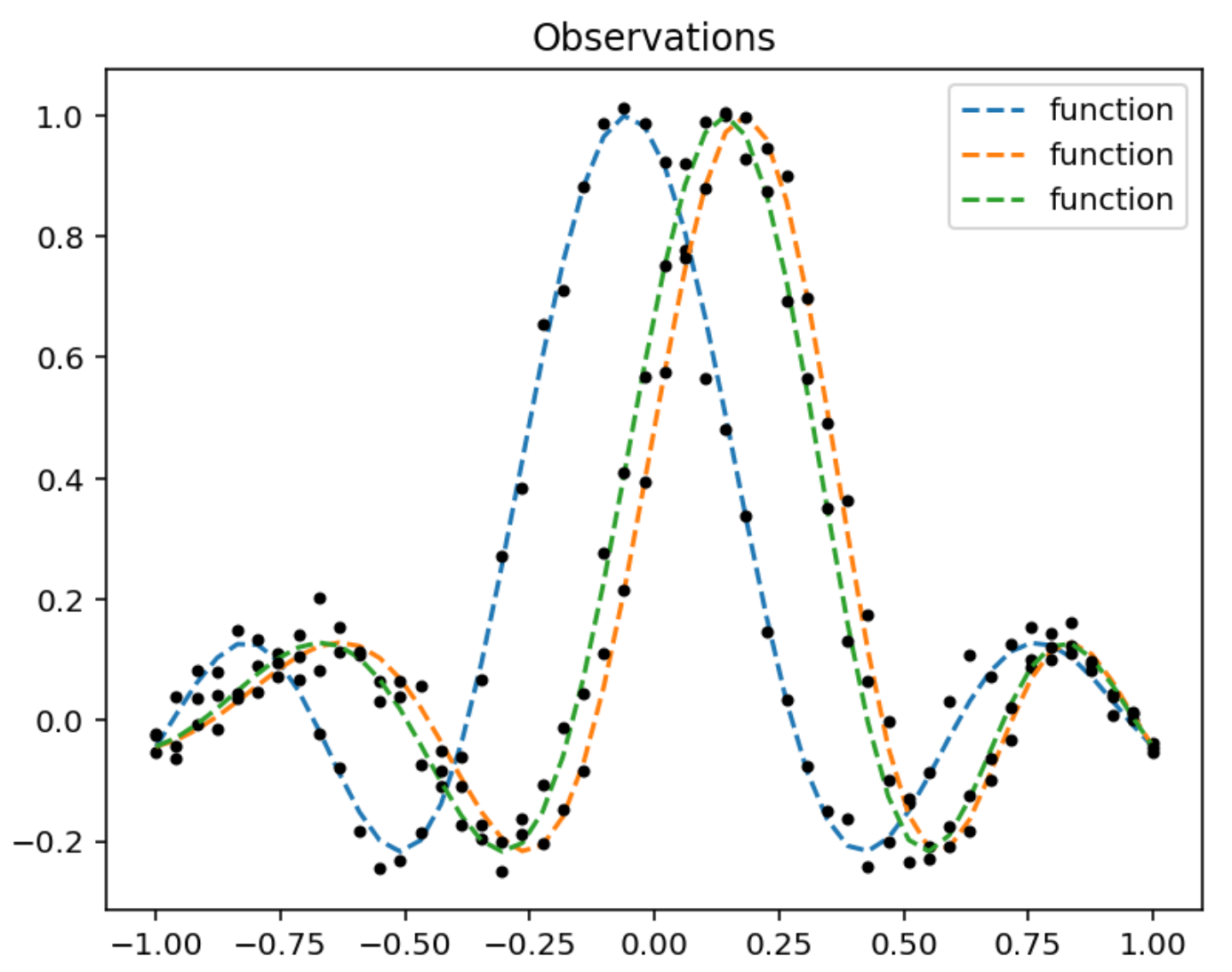}}
      \vskip1em
      \subcaptionbox{Examples of sampled aligned functions $\s$.\label{fig:supp:align:c}}{\includegraphics[width=\hsize]{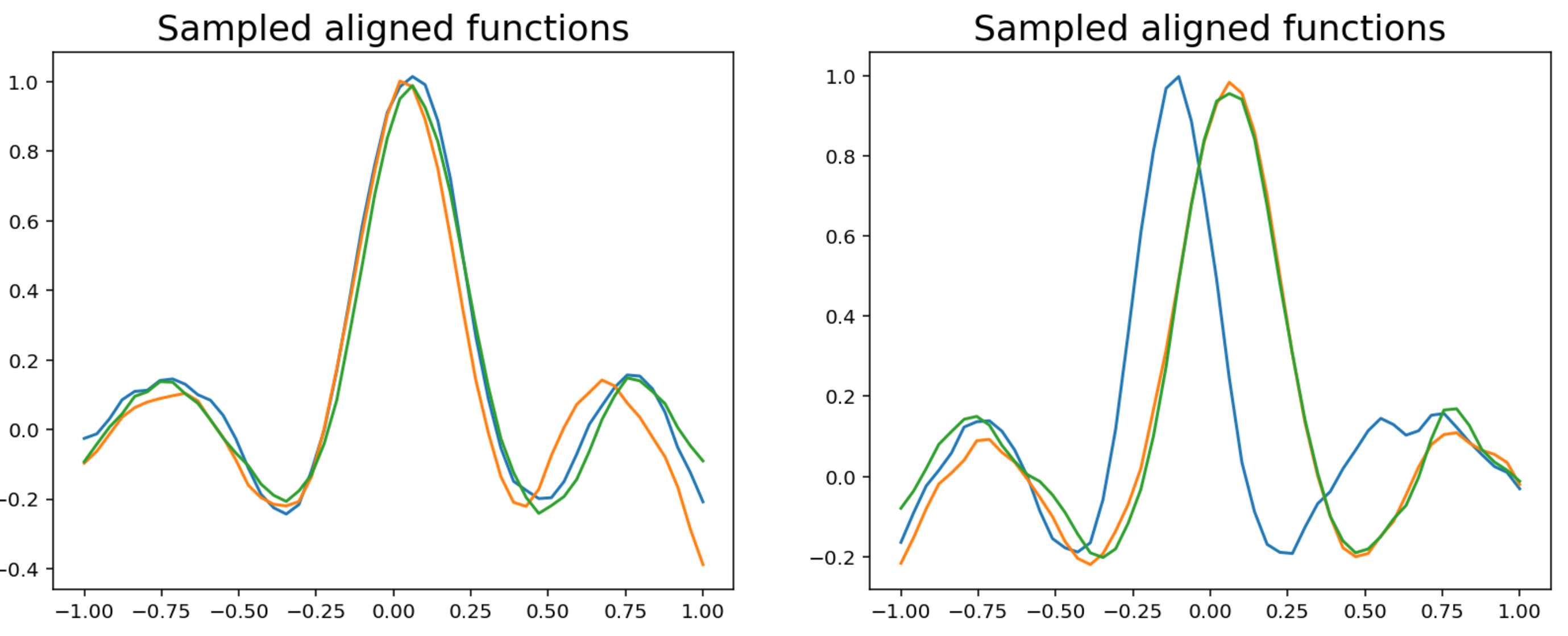}}  
    }
    \hfill%\hskip1em
    \parbox{.6\figrasterwd}{\centering%
      \subcaptionbox{Fitted sequences (left), estimated warps (middle) and fits in the warped coordinates (right) for the 3 sequences.}{\includegraphics[width=\hsize, height=8cm]{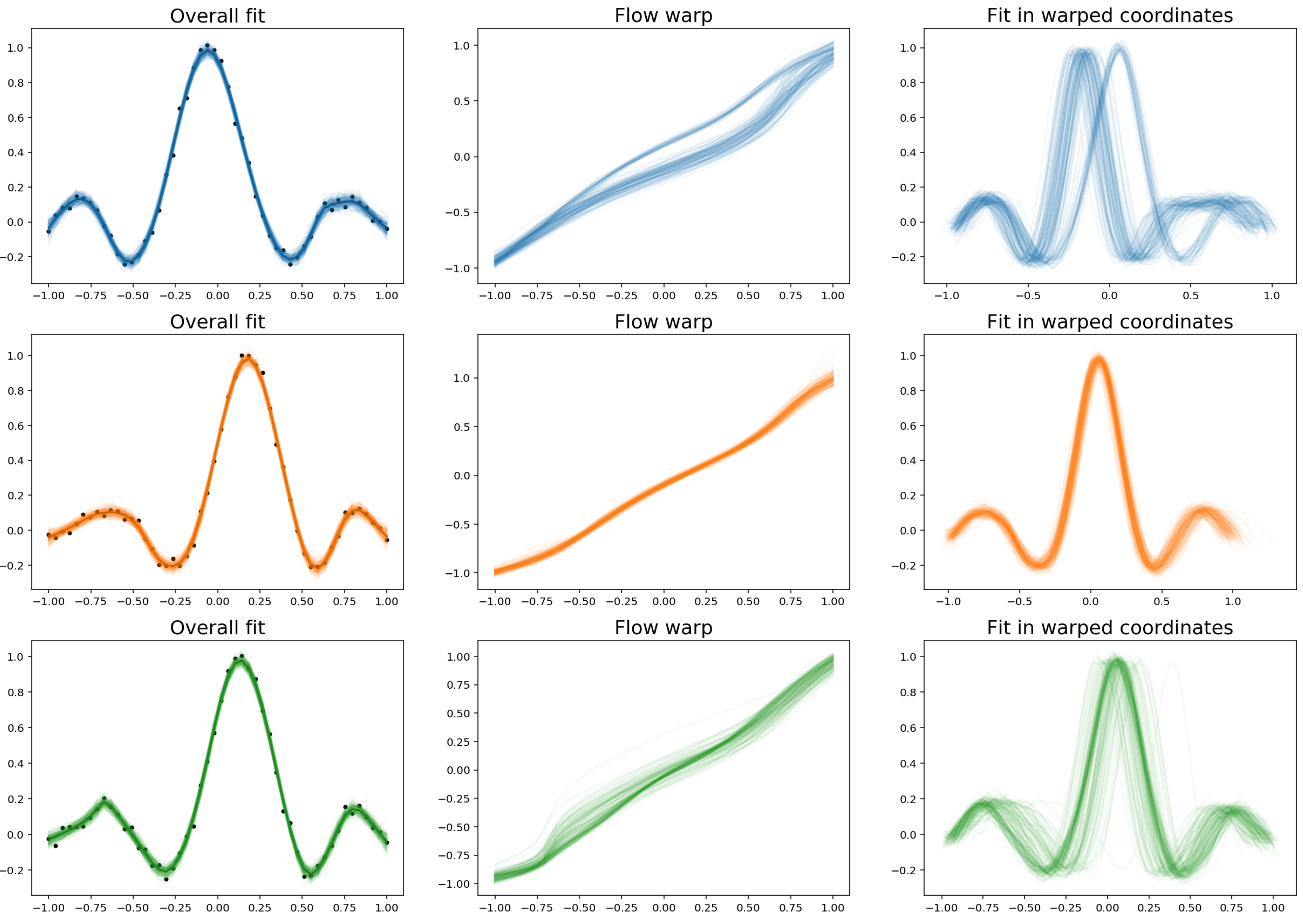}}%'height=8cm' is needed for this example only and can be dropped when using it with actual images
    }
  }
    \caption{Illustration of uncertainty in warps and cluster assignments. When the warps and the cluster assignment are allowed to be bi-modal, and model captures two possible solutions, one that assigns all sequences to a single cluster and aligns them within the cluster, and another solution that favours the model with two separate clusters. This can be see in the fit in warped coordinates figure for the blue curve where the majority of the samples are assigned to one cluster (which corresponds to now aligning the blue function to the other, as shown on the right in Fig.~\ref{fig:supp:align:c}) while a small subset is assigned to a new cluster (which corresponds to all sequences being aligned together, as shown on the left in Fig.~\ref{fig:supp:align:c}). }
    \label{fig:supp:align}
\end{figure*}

\subsection{Quantitative results}
\label{supp:quantitative-results}

The expected log posterior predictive density is an evaluation metric defined as:
\begin{equation}
\begin{aligned}
    \text{ELPD} &= \log \int p(y^* \given f^*) p(f^* \given \y) \text{d} f^* \\
    & \approx  \log \int p(y^* \given f^*) q(f^* \given \y) \text{d} f^*.
\end{aligned}
\end{equation} The results on the data described in Sec.~5 (with 100 data points) for the GP with derivatives~\cite{Riihimaki:2010}, the transformed GP~\cite{Andersen:2018} and the monotonic flow are given in Table~\ref{table:elpd}.

% !TEX root = ../neurips_2019.tex
\begin{table*}[]
\caption{Root-mean-square error $\pm$ SD ($\times 100$) of $20$ trials for data of size $N = 100$}
\label{table:evals}
\begin{adjustbox}{center}
\resizebox{0.9\textwidth}{!}{%
\begin{tabular}{llllllll}
\toprule
                                                                                                  & flat       & sinusoidal  & step        & linear      & exponential & logistic    \\
\midrule
GP                                                                                                & 15.1       & 21.9        & 27.1        & 16.7        & 19.7        & 25.5        \\
GP projection~\cite{Lin:2014}                                               & 11.3       & 21.1        & 25.3        & 16.3        & 19.1        & 22.4        \\
Regression splines~\cite{Shively:2009}                                      & \phantom{1}9.7        & 22.9        & 28.5        & 24.0          & 21.3        & 19.4        \\
GP approximation~\cite{Maatouk:2017}                                        & \phantom{1}8.2        & 20.6        & 41.1        & 15.8        & 20.8        & 21.0          \\
GP with derivatives~\cite{Riihimaki:2010}                        & 16.5 $\pm$ 5.1 & 19.9 $\pm$  2.9 & 68.6 $\pm$ 5.5 & 16.3 $\pm$  7.6 & 27.4 $\pm$ 6.5 & 30.1 $\pm$ 5.7         \\
Transformed GP~\cite{Andersen:2018} {\tiny (VI-full)} & \phantom{1}\textbf{6.4} $\pm$ 4.5 & 20.6 $\pm$ 5.9 & 52.5 $\pm$ 3.6 & \textbf{11.6} $\pm$ 5.8 & 17.5 $\pm$ 7.3 & \textbf{17.1} $\pm$ 6.2 \\
\textbf{Monotonic Flow (ours)}                                   & \phantom{1}6.8 $\pm$ 3.2 & \textbf{17.9} $\pm$ 4.2 & \textbf{20.5} $\pm$ 5.0 & 13.2 $\pm$ 6.7 & \textbf{14.4} $\pm$ 4.8 & 18.1 $\pm$ 5.0 \\
\bottomrule
\end{tabular}
}%resizebox
\end{adjustbox}
\end{table*}
% !TEX root = ../neurips_2019.tex

\begin{table*}[]
\caption{Root-mean-square error $\pm$ SD ($\times 100$) of 20 trials for data of size $N = 15$}
\label{tab:15pts}
\begin{adjustbox}{center}
\resizebox{0.9\textwidth}{!}{%
\begin{tabular}{lllllll}
\toprule
        & flat            & sinusoidal      & step             & linear          & exponential  & logistic   \\
\midrule
Transformed GP~\cite{Andersen:2018} {\tiny (VI-full)} & \textbf{18.5} $\pm$ 14.4 & 40.0 $\pm$ 17.5 & 101.9 $\pm$ 11.4 & 37.4 $\pm$ 22.8 & 52.9 $\pm$ 11.9 & 51.7 $\pm$ 19.6 \\
\textbf{Monotonic Flow (ours)}    & 21.7 $\pm$ 15.0 & \textbf{39.1} $\pm$ 13.0 & \phantom{1}\textbf{64.5} $\pm$ 10.7  & \textbf{30.8} $\pm$ 12.0 & \textbf{32.8} $\pm$ 17.9 & \textbf{43.2} $\pm$ 15.2 \\
\bottomrule
\end{tabular}
}%resizebox
\end{adjustbox}
\end{table*}
% !TEX root = ../neurips_2019.tex
\begin{table*}[]
\caption{Expected log posterior predictive density estimate ($\pm$ SD) of $20$ trials for data of size $N = 100$}
\label{table:elpd}
\begin{adjustbox}{center}
\resizebox{0.9\textwidth}{!}{%
\begin{tabular}{llllllll}
\toprule
          & flat       & sinusoidal  & step        & linear      & exponential & logistic    \\
\midrule
GP with derivatives~\cite{Riihimaki:2010}   & -1.43 $\pm$ 0.08 & -1.41 $\pm$  0.06 & -1.69 $\pm$ 0.15 & -1.36 $\pm$ 0.04 & -1.45 $\pm$ 0.08 & -1.45 $\pm$ 0.11         \\
Transformed GP~\cite{Andersen:2018} {\tiny (VI-full)} & -1.44 $\pm$ 0.03 & -1.39 $\pm$ 0.02 & -1.51 $\pm$ 0.06 & -1.40 $\pm$ 0.03 & -1.41 $\pm$ 0.02 & -1.41 $\pm$ 0.02 \\
\textbf{Monotonic Flow (ours)}  & -1.39 $\pm$ 0.05 & -1.42 $\pm$ 0.05 & -1.41 $\pm$ 0.08 & -1.39 $\pm$ 0.05 & -1.40 $\pm$ 0.07 & -1.43 $\pm$ 0.07 \\
\bottomrule
\end{tabular}
}%resizebox
\end{adjustbox}
\end{table*}

% \begin{table*}[]
% \caption{Expected log posterior predictive density estimate ($\pm$ SD) of $20$ trials for data of size $n = 100$}
% \label{table:elpd}
% \begin{adjustbox}{center}
% \resizebox{0.95\textwidth}{!}{%
% \begin{tabular}{llllllll}
% \toprule
%           & flat       & sinusoidal  & step        & linear      & exponential & logistic    \\
% \midrule
% GP with derivatives~\cite{Riihimaki:2010}   & -1.433 $\pm$ 0.078 & -1.412 $\pm$  0.062 & -1.692 $\pm$ 0.150 & -1.360 $\pm$ 0.035 & -1.446 $\pm$ 0.075 & -1.450 $\pm$ 0.106         \\
% Transformed GP~\cite{Andersen:2018} {\tiny (VI-full)} & -1.439 $\pm$ 0.025 & -1.392 $\pm$ 0.020 & -1.512 $\pm$ 0.055 & -1.404 $\pm$ 0.030 & -1.407 $\pm$ 0.020 & -1.413 $\pm$ 0.018 \\
% \textbf{Monotonic Flow (ours)}  & -1.387 $\pm$ 0.054 & -1.419 $\pm$ 0.052 & -1.407 $\pm$ 0.083 & -1.393 $\pm$ 0.054 & -1.402 $\pm$ 0.073 & -1.427 $\pm$ 0.065 \\
% \bottomrule
% \end{tabular}
% }%resizebox
% \end{adjustbox}
% \end{table*}

\begin{figure*}[]
\centering
    \begin{subfigure}[t]{0.32\textwidth}%{0.47\textwidth}
        \includegraphics[width=\textwidth]{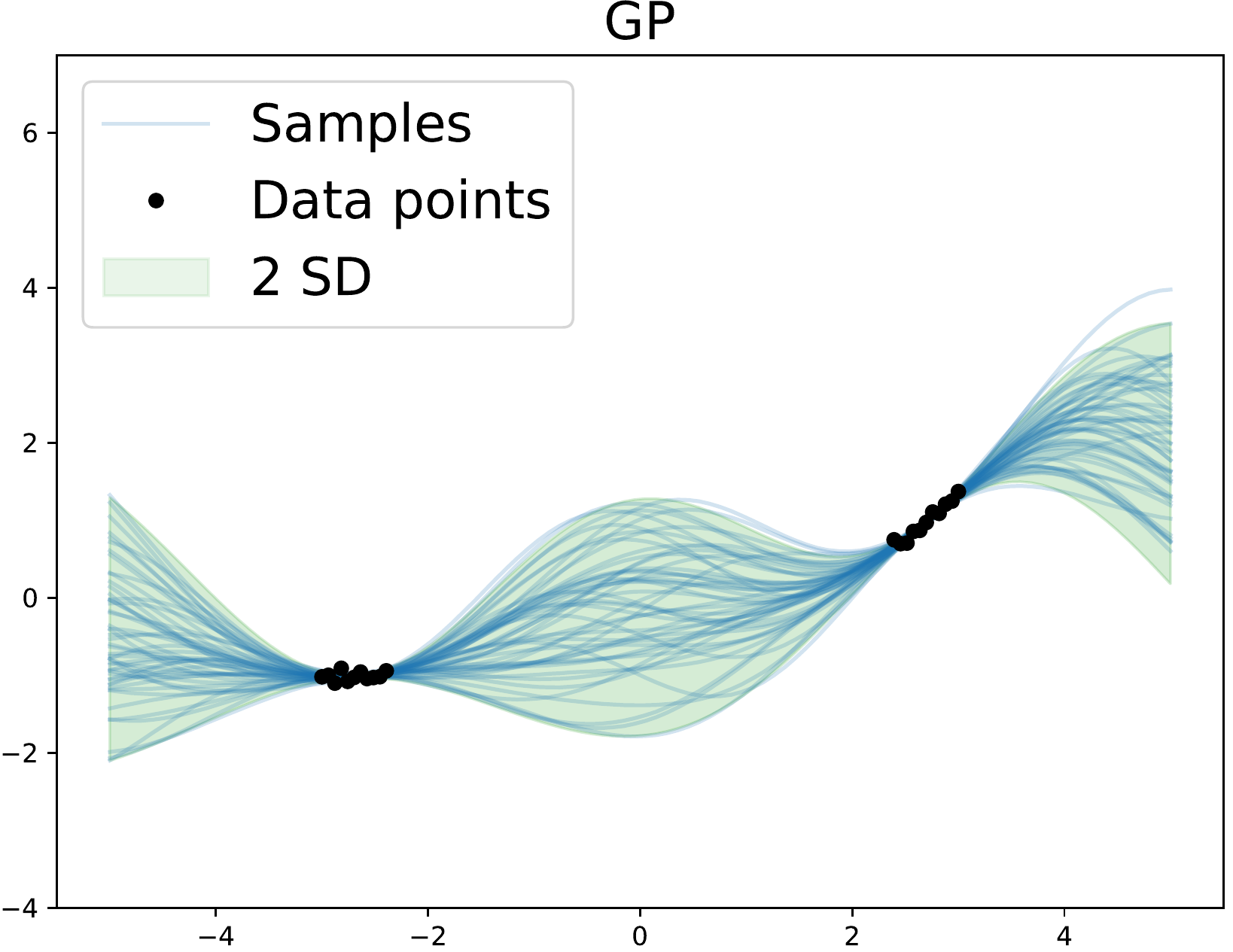}
        \caption{Standard GP.}\label{fig:CI:1}
    \end{subfigure} \hfill%\hspace{4pt}
    \begin{subfigure}[t]{0.32\textwidth}
        \includegraphics[width=\textwidth]{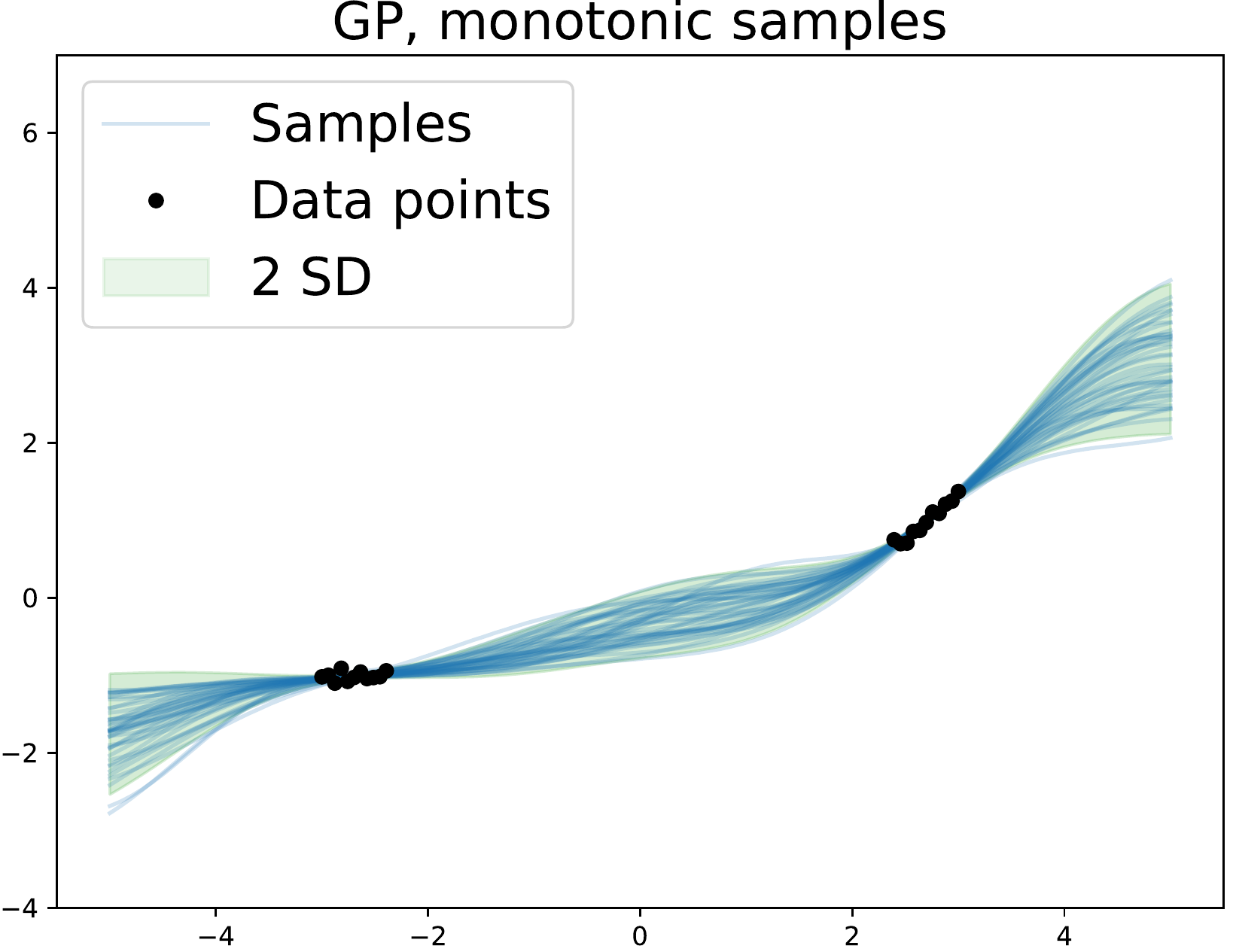}
        \caption{GP, monotonic samples.} \label{fig:CI:2}
    \end{subfigure}  \hfill%\\ \vspace{0.35cm}
    \begin{subfigure}[t]{0.32\textwidth}
        \includegraphics[width=\textwidth]{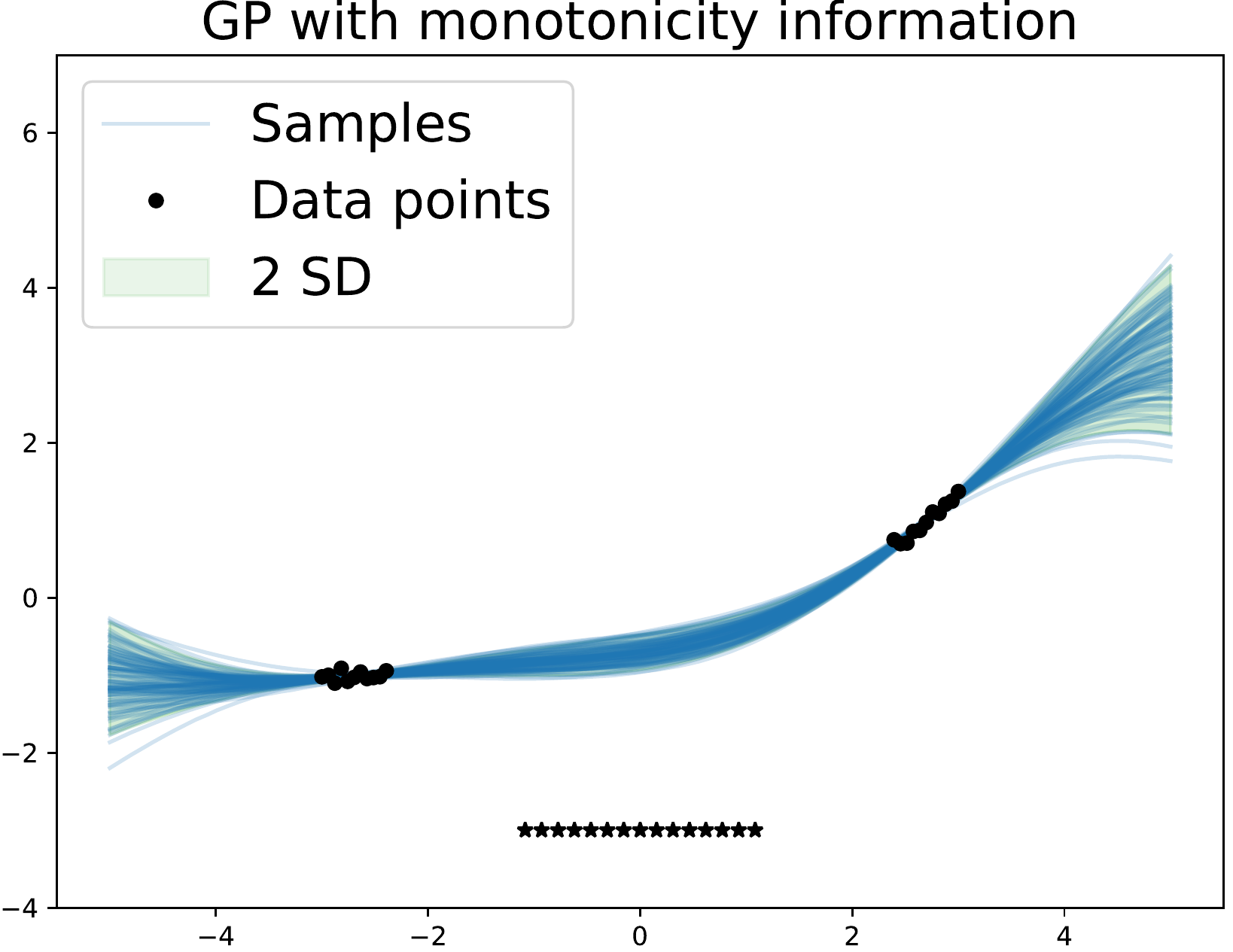}
        \caption{GP with monotonic information.}\label{fig:CI:3}
    \end{subfigure} \\\vspace{0.35cm}%\hspace{4pt}
    \begin{subfigure}[t]{0.32\textwidth}
        \includegraphics[width=\textwidth]{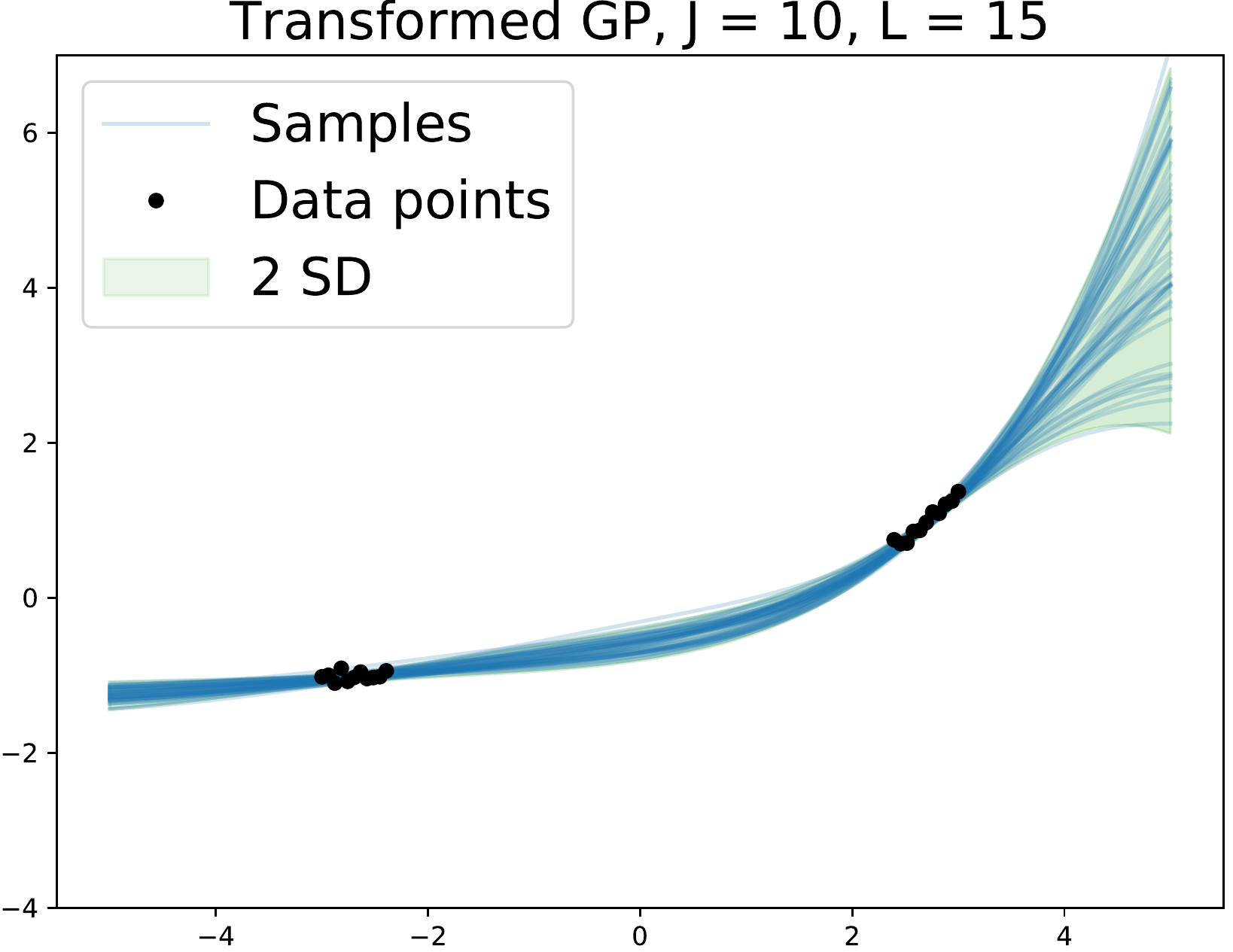}
                \caption{Transformed GP (VI-full).}\label{fig:CI:4}
    \end{subfigure} \hspace{20pt}%\\ \vspace{0.35cm}
    \begin{subfigure}[t]{0.32\textwidth}
        \includegraphics[width=\textwidth]{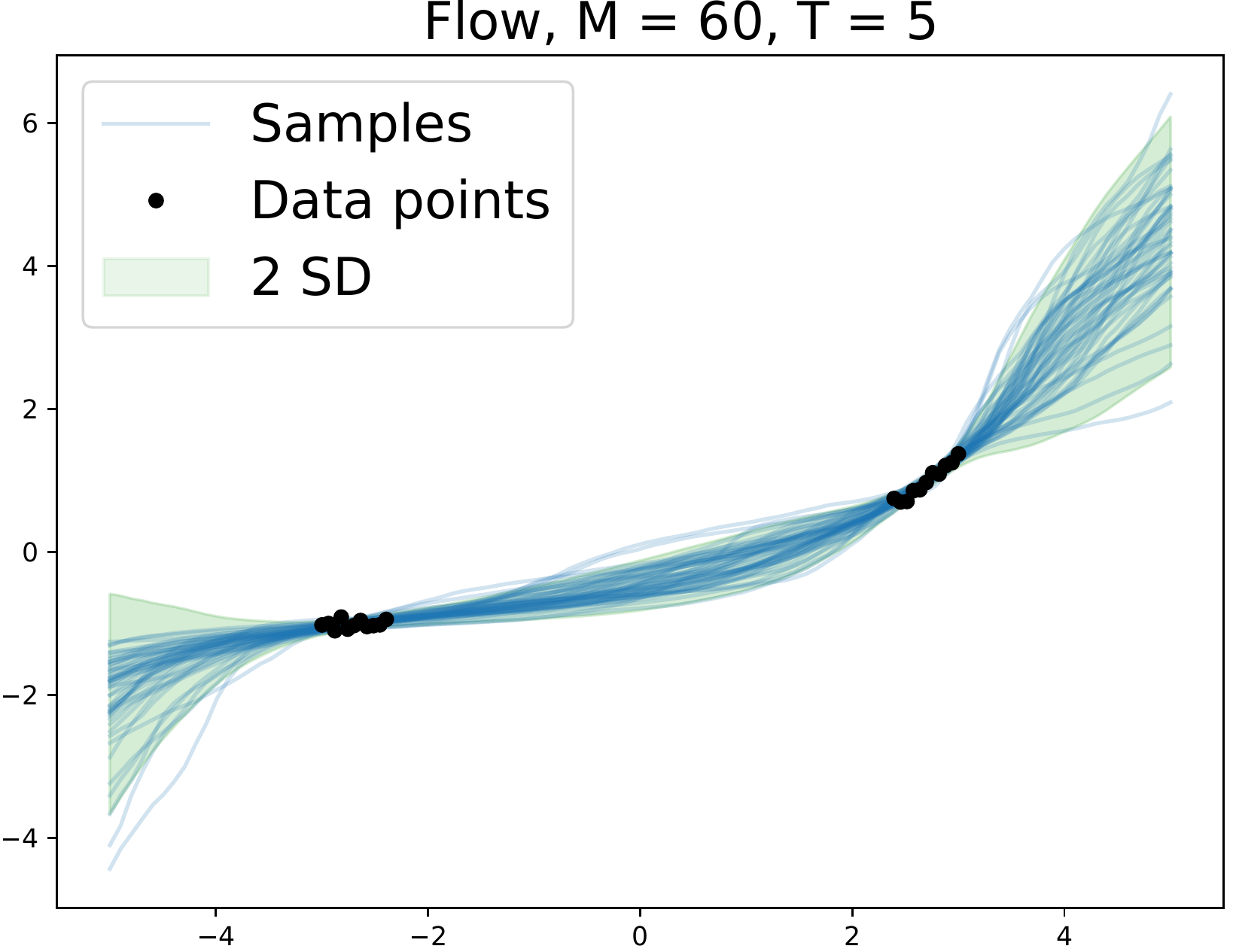}
        \caption{Flow (ours).}\label{fig:CI:5}
    \end{subfigure} 
    \caption{Comparison of the confidence intervals for standard GP, and monotonic regression methods (GP with monotonic information from~\cite{Riihimaki:2010} and Transformed GP from~\cite{Andersen:2018}). The samples from the fitted models are shown in blue and the 2 standard deviations from the mean are shown in green.} \label{fig:confidence-intervals}    
\end{figure*}

% \newpage
% \bibliography{references}
% %\bibliographystyle{icml2018}
% \bibliographystyle{plain} % {apalike} Guidelines want a full author name + year

\end{document}